\documentclass[11pt]{article}

\PassOptionsToPackage{dvipsnames}{xcolor}

\usepackage[final]{acl}

\usepackage{times}
\usepackage{latexsym}

\usepackage[T1]{fontenc}

\usepackage[utf8]{inputenc}

\usepackage{microtype}

\usepackage{inconsolata}

\usepackage{graphicx}
\usepackage{hyperref}

\usepackage{amsmath}
\usepackage{amssymb}
\usepackage{mathtools}
\usepackage{amsthm}

\usepackage[capitalize,noabbrev]{cleveref}

\usepackage{xcolor}
\usepackage{pgfplots}
\usepackage{listings}
\usepackage{soul}
\usepackage[export]{adjustbox}

\usepackage{tabularx}
\usepackage{algorithm}
\usepackage{multirow}
\usepackage{makecell}
\usepackage{subcaption}
\usepackage{fontawesome}
\usepackage[shortlabels]{enumitem}
\usepackage{booktabs}
\usepackage{xspace}

\newsavebox{\largestimage}

\lstdefinestyle{promptstyle}{
    basicstyle=\small\ttfamily,
    breaklines=true,
    breakatwhitespace=true,
    frame=single,
    backgroundcolor=\color{gray!10},
    columns=fullflexible,
    keepspaces=true
}

\theoremstyle{plain}

\theoremstyle{definition}

\theoremstyle{remark}

\usepackage[textsize=tiny]{todonotes}

\usepackage{environ}
\NewEnviron{smallalign}{%
  \par
  \begingroup\small
  \abovedisplayskip=6pt
  \belowdisplayskip=6pt
  \abovedisplayshortskip=2pt
  \belowdisplayshortskip=2pt
  \begin{align}\BODY\end{align}%
  \endgroup
}

\begin{document}

\title{Intra-Prompt Parallel Decoding for Common-Context Question Answering}

\author{
  Theodore Glavas\thanks{\hspace{1mm}This work was done as part of an internship at Amazon.}$^{1,2,3,4}$ \quad Nikhita Vedula$^{1}$ \quad Dushyanta Dhyani$^{1}$ \\
  \textbf{Antonios Valkanas}$^{*1,2,3,4}$ \quad \textbf{Yilun Zhu}\thanks{\hspace{1mm}Work done while at Amazon. Currently at Apple.}$^{1}$ \quad \textbf{Shervin Malmasi}$^{1}$ \\
  \vspace{2pt} \\
  $^{1}$ Amazon.com, Inc. \quad $^{2}$ McGill University \quad $^{3}$ Mila \quad 
$^{4}$ Int. Lab. Learning Systems
\\
 \small{\texttt{\{theodore.glavas, antonios.valkanas\}@mail.mcgill.ca, yz565@georgetown.edu,}}\\ \small{\texttt{\{veduln, dhyanidd, malmasi\}@amazon.com}}
}

\maketitle

\begin{abstract}
 In common-context question answering (CCQA) tasks, multiple input questions share a common context to base their answers from. However, Large Language Models typically generate each answer using an independent prompt. While existing batching and caching techniques help improve parallelism and reduce repeated computations, the separation of questions across prompts limits the achievable speedup, as modern GPUs are underutilized due to a memory bottleneck during attention. We present Intra-Prompt Parallel Decoding (IPPD), a novel inference method that answers multiple common-context questions in parallel within a single prompt. IPPD directly addresses the bottleneck by efficiently sharing both memory and computation during the attention process, as the next token for every question is decoded in a single inference step. IPPD uses virtual position IDs and attention mask manipulation to generate the same output as standard prompting without requiring fine-tuning or any changes to the LLM architecture. Since all parallelism occurs within a prompt, IPPD is fully compatible with batched inference, even when each prompt features a different context. Our experiments show that IPPD delivers up to 7X the effective throughput as standard decoding without quality degradation, and outperforms prefix caching with PagedAttention in most settings.
\end{abstract}

\section{Introduction}
Text generation has emerged as a core capability of LLMs, powering downstream NLP applications such as dialogue systems \citep{yi2025surveyrecentadvancesllmbased}, summarization \citep{zhang2025comprehensivesurveyprocessorientedautomatic}, and question answering (QA) \citep{yue2025surveylargelanguagemodel}. Within the broad landscape of text generation, contextual QA stands out as a representative benchmark task: given a document and a question, a model must not only locate relevant pieces of evidence within the context, but also integrate and reason across them to answer the question. A large subset of QA tasks referred to as \textit{reading comprehension} follows this setting, where multiple questions are asked about the same passage \citep{lai2017_race, yang-etal-2018-hotpotqa, rajpurkar2018_squadv2, kocisky2018_narrativeqa}. 
We define \textbf{common-context question answering (CCQA)} as the task of producing multiple outputs by answering different questions about a shared context, such as a document. This setting is common in industrial information extraction, where systems must extract numerous fields or facts from the same long, unstructured input, including financial statements, medical records, legal documents, and web pages \citep{IDPLeaderboard, Brinkmann2024_ExtractGPT}.

Although LLMs excel at generating free-form answers for a wide range of applications, the computational cost of generative LLMs limits their adoption in large-scale real-world scenarios. This challenge is particularly critical in CCQA because, despite substantial context overlap across questions, autoregressive decoding repeatedly computes the same prompt and passage for each question, wasting GPU resources with redundant computations and amplifying inefficiencies.

Motivated by cost savings, a significant body of literature has focused on improving the efficiency of LLM inference for text generation as well as CCQA. A major performance bottleneck arises from the challenge of fully utilizing highly-parallel GPU architectures, due to the high memory requirements and sequential nature of autoregressive decoding. State-of-the-art serving platforms, such as vLLM \citep{kwon2023efficient}, implement a variety of advanced methods to address this. Continuous batching dynamically groups incoming requests to maximize GPU utilization, PagedAttention \citep{kwon2023efficient} shares the memory-intensive key-value (KV) cache across requests, and Cascade inference \citep{cascade-inference,juravsky2024hydragen} shares the compute-expensive attention score calculation of shared prefixes across requests. While these innovations have substantially accelerated inference, the computational cost of processing large, common contexts for millions of queries remains a key challenge, particularly in offline scenarios requiring high throughput, and when not all batched requests share the same context.

\begin{figure*}[ht!]
  \centering
  \includegraphics[width=0.75\textwidth]{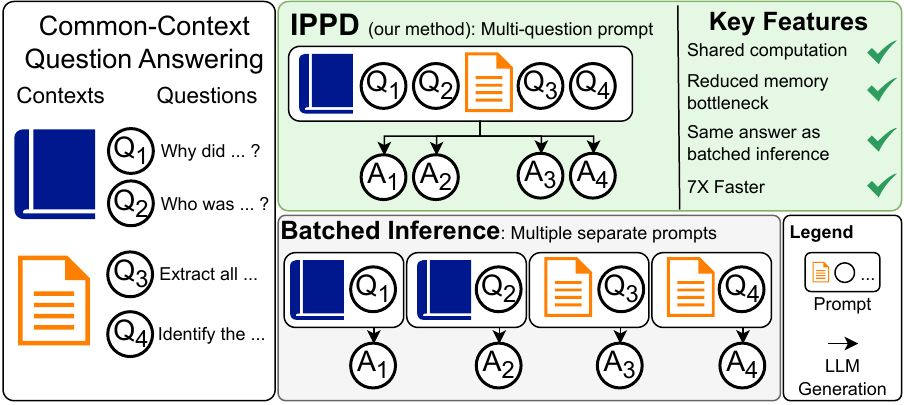}
  \caption{In QA tasks, multiple questions often reference the same context. Traditional batched inference separates each question into a separate prompt, and processes prompts independently to generate the answer. IPPD combines multiple questions and contexts within a single prompt and jointly decodes multiple answers in parallel. This eliminates repeated computations and reduces the memory bottleneck of GPUs during attention to generate the same answers in a fraction of the time.} %
  \label{fig:figure1}
\end{figure*}

To address the above challenges, we introduce Intra-Prompt Parallel Decoding (IPPD),\footnote{Code is available at \href{https://github.com/networkslab/IPPD}{https://github.com/networkslab/IPPD} Corresponding author: \texttt{theodore.glavas@mail.mcgill.ca}} a novel inference method for maximizing the batched throughput of CCQA with intra-prompt parallelism (\Cref{fig:figure1}), where multiple questions are combined into a single prompt, and answers are decoded in parallel within it. By decoding tokens out of order and adjusting the attention mask, IPPD increases throughput while not affecting the final output. The increased efficiency of IPPD stems from dramatically reducing the number of required memory accesses during attention, which is known to be heavily memory bottlenecked \citep{Recasens2025MindTheGap}. Importantly, IPPD can be easily applied to existing systems without changes to the core model implementation, maintains compatibility with orthogonal inference acceleration methods, and is up to 7 times more computationally efficient than standard batched inference on realistic CCQA tasks.

\section{Related Work}
\label{sec:related_work}
\paragraph{Token parallelism} 
Recent efforts to accelerate LLM inference have explored decoding multiple tokens in parallel within a prompt. A common approach is to decode consecutive tokens in parallel. Speculative decoding \citep{leviathan2023_spec_decoding} employs a smaller draft model to propose a sequence of tokens to be verified in parallel by a larger verifier LLM. The propose-then-verify paradigm is also utilized in \citet{Cai2024_medusa} and \citet{Fu2024_lookahead_decoding}. A key limitation of these approaches is that the length of the drafted token sequence is directly tied to the level of parallelism, and longer drafts are more prone to quality degradation and subsequent rejection by the verifier model. In contrast, our method parallelizes non-consecutive tokens known to be independent, which eliminates the need for a verification step. Other work has explored parallelizing non-sequential tokens. \citet{ning2024_skeletonofthought} and \citet{liu2024_APAR} break down complex tasks into independent sub-tasks that can be processed in parallel. While this can reduce single-task latency, it requires creating and processing multiple new prompts. We show in \Cref{sec:comp_efficiency} that intra-prompt parallelism results in denser computations during attention that more effectively utilize modern GPU hardware compared to batching separate prompts, resulting in higher throughput in offline settings.

\paragraph{Batch parallelism}
Another relevant research direction is accelerating LLM inference by sharing memory and computation across batched prompts. Prefix caching \citep{kwon2023efficient, ye-etal-2024-chunkattention, pan2025marconiprefixcachingera} and PagedAttention \citep{kwon2023efficient} are common methods that create a shared Key-Value (KV) cache to reduce memory usage and avoid recomputing KV matrices for common prefixes in batched prompts. While these techniques reduce the required floating point operations (FLOPs) during attention, they are limited by the large number of memory accesses required to independently compute attention scores for each prompt, resulting in GPU underutilization. We elaborate on this phenomenon in \Cref{sec:comp_efficiency}.

\paragraph{Tree-based decoding}
Recent work has demonstrated the success of using tree data structures to share memory and computation of partially overlapping queries. A KV cache tree stores shared contexts without duplication, and cross-attention is performed in chunks between the incoming queries and shared KV \citep{cascade-inference,juravsky2024hydragen,yao2025deft,wang2025flashforgeultraefficientprefixawareattention}. The final attention scores are then computed using a merge operator on the partial attention scores. Although strong performance can be obtained by designing optimized low-level attention kernels, this approach suffers from the overhead of the merge operation. Since a custom kernel and KV cache manager are required, existing methods are often limited in their batching functionality, or incompatible with newer models and inference platforms \citep{yao2025deft,pan2025fasttree,cascade-inference}. These factors limit their applicability and longevity. Rather than breaking up the attention process into many parts, IPPD combines common-context questions into a single KV cache and attention operation without any context duplication. By supplying the model with a custom attention mask, IPPD parallelizes CCQA tasks without modifying the model implementation or breaking compatibility with orthogonal methods.

\section{Problem Formulation: CCQA} \label{sec:problem_setting}
We begin by denoting the inputs to the LLM in a QA setting. All LLM inputs are comprised of the following components: a \emph{common instruction} or prompt prefix $p$, a \emph{context} $c$ (for example a document or passage), and a \emph{question} $x$. Usually, these three logical parts form a triplet $(p,c,x)$ which, after tokenization, becomes a contiguous sequence presented to the model. In realistic workflows, we can have many such triplets, and several triplets may share the same context. To capture this, let $\{c_j\}_{j=1}^J$ denote contexts and for each context $c_j$ let $\{x_{j,k}\}_{k=1}^{M_j}$ denote the associated questions. The set of logical triplets is then $\mathcal{T}=\{(p,c_j,x_{j,k}) \mid j=1,\dots,J,\; k=1,\dots,M_j\}$. We assume this set to be fully available offline at inference time. We further assume that each question can be answered independently: generating $\hat{y}_{j,k}$ may depend on $p$, $c_j$, $x_{j,k}$, and its own previously generated tokens, but not on the answers to other questions.

A traditional LLM loop to process the triplets $\mathcal{T}$ would be to issue $|\mathcal{T}|$ independent prompts, each requiring a separate LLM forward pass $f(\cdot)$. 
During autoregressive answer generation, each answer token is generated and fed back into the input sequence and passed through forward pass $f$. As a result, the $t$-th token of the answer is obtained using $\hat{y}_{j,k,t} = f(p, c_j, x_{j,k}, y_{j,k,<t})$, where $y_{j,k,<t}$ represents the first $t-1$ answer tokens generated for question index $k$ of context $c_j$.
At the end of the process, we obtain the answer set $\mathcal{A} = \{ \hat{y}_{j,k} \mid j=1,\dots,J,\; k=1,\dots,M_j\}$. 

Traditional inference methods employ batching to process multiple prompts concurrently. Although parts of the computation, such as the MLP layers, can be effectively parallelized across prompts, computations in the attention layers are performed independently for each prompt. This introduces a bottleneck for batched inference, particularly when batch elements share a common context $c_j$. We elaborate on this phenomenon in \cref{sec:methodology}. To overcome this limitation, we wish to develop an inference method that can efficiently share computation in the attention layers, while maintaining batched inference compatibility and not requiring any training or fine-tuning. To achieve this, we aim to parallelize the generation of $\hat{y}_{j,k}$ across $j, k$ \emph{within} a prompt. It is important to note that each token \emph{within an answer} $\hat{y}_{j,k,t}$ will still be generated autoregressively, as we are parallelizing across the \textit{question set}. The core methodological challenges are ensuring that the introduced intra-prompt parallelism is computationally efficient and does not affect the quality of the generated answers $\mathcal{A}$.

We evaluate candidate strategies by comparing against batched autoregressive generation along two complementary axes: \textit{throughput} (efficiency) and \textit{answer quality} (fidelity to the ground truth question answers). Throughput measures how many logical triplets are answered per unit time per GPU. Answer quality is measured at the task level using standard metrics (Accuracy, F1, ROUGE-L). Task-level metrics compare generated answers $\hat y_{j,k}$ to ground truth answers $y_{j,k}$. See Appendix~\ref{app:exp_details} for definitions of throughput and metrics.

\begin{figure*}[ht!]
  \centering
  \includegraphics[width=0.90\textwidth,height=8.5cm,keepaspectratio]{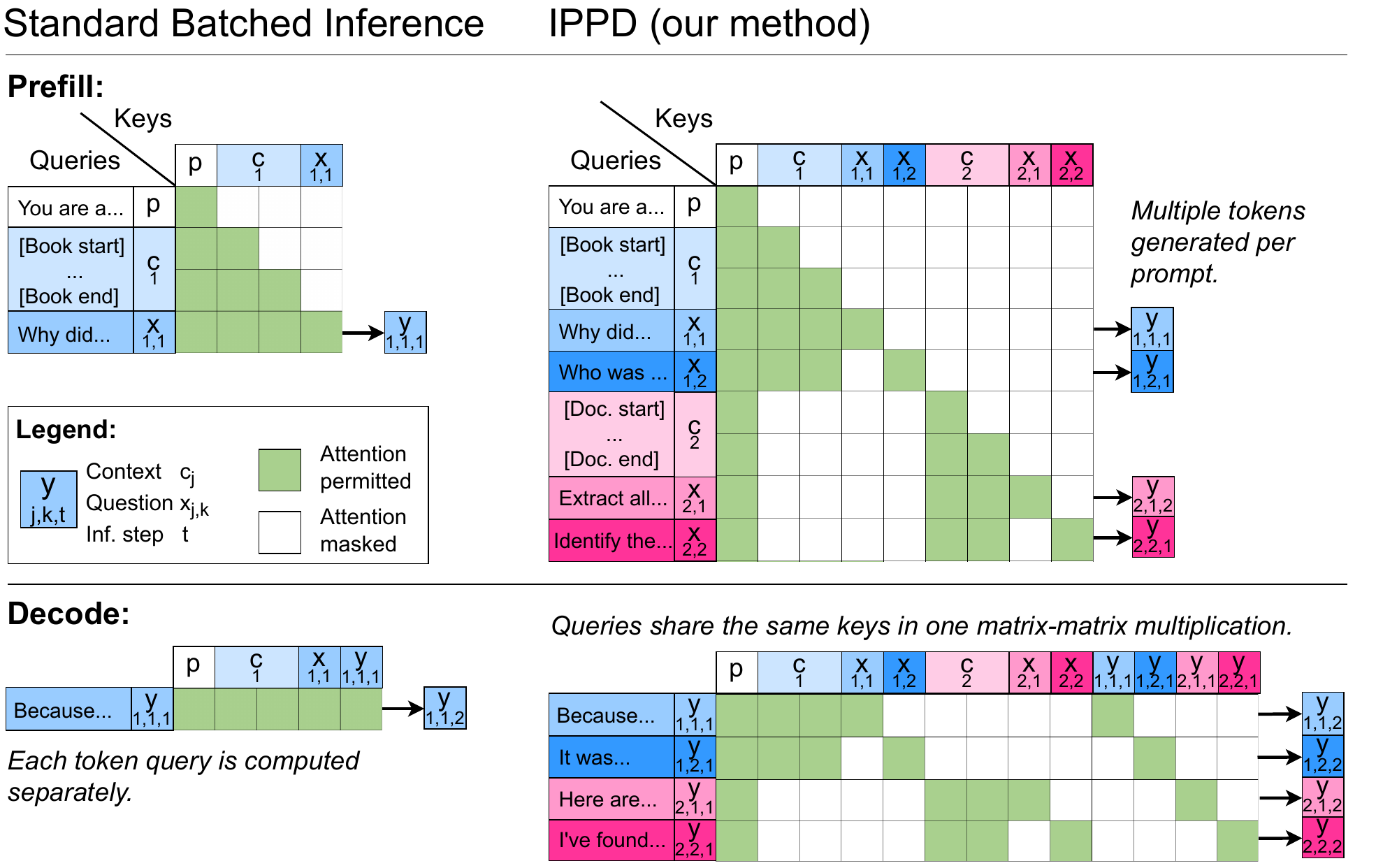}
  \caption{The attention process between the queries (rows) and keys (columns) is depicted with green cells representing unmasked query-key pairs. In standard batched inference, the sequence contains a common instruction $p$, shared context $c$, question $x$ and partial answer $y$ (simplified as one or two tokens for illustrative purposes). The arrows indicate which token is generated using the output of that query row. IPPD concatenates multiple contexts and queries into a single stacked prompt. The attention mask ensures that each query can only attend to past tokens from the relevant context and question, identifiable by color and shade respectively. At inference step $t$, IPPD generates the $t$-th token of four answers through a single shared attention operation per layer in this example. Batched inference would require four independent attention calculations to achieve the same result.}
  \label{fig:figure2}
\end{figure*}

\section{Method: Intra-Prompt Parallel Decoding (IPPD)} \label{sec:methodology}
We introduce \textit{Intra-Prompt Parallel Decoding (IPPD)}, a strategy that enables multiple context–question triplets to be processed in parallel within a single prompt. \Cref{sec:pre-processing} describes how we replace the $|\mathcal{T}|$ independent forward passes of the canonical autoregressive loop with a single, structured forward-pass strategy that: (i) stacks contexts and questions into one prompt, (ii) assigns hierarchical position identifiers so positional relationships do not induce cross-question leakage, (iii) builds an attention mask that enforces per-question causal inference, and (iv) decodes tokens in parallel across the question set. \Cref{fig:figure2} provides an illustrative example of the process: the IPPD prompt combines two contexts $c_1$ and $c_2$ with two questions each, $x_{1,1},x_{1,2}$ and $x_{2,1},x_{2,2}$ respectively. During the prefill stage, the attention mask prevents the questions from (a) attending to each other, and (b) attending to the non-relevant context. After a full forward pass, we can extract four answer tokens, corresponding to the first token of each question. These tokens are passed in parallel during the decode stage with appropriate masking to maintain high parallelism throughout generation. Our goal is to ensure that each generated token's conditional distribution under this multi-query, per-step parallel scheme matches the distribution produced by independent autoregressive decoding on the corresponding triplet $(p,c_j,x_{j,k})$, which we demonstrate in \Cref{sec:parallel-decoding}. \Cref{sec:comp_efficiency} describes the computational efficiency of IPPD, arising from increasing GPU utilization during attention.

\subsection{Input pre-processing} \label{sec:pre-processing}
We construct a stacked token sequence $s$ containing the global instruction $p$, $\ell\leq J$ contexts, and the associated questions. We append the answers \emph{after} this header. The header region of the prompt is 
\begin{smallalign}
    \big[\, p,\; c_1,\; x_{1,1},\; x_{1,2},\; \dots,\; c_2,\; x_{2,1},\; \dots, c_\ell,\; x_{\ell,1},\;\,\dots],
\end{smallalign}
and the full token sequence $s$ at any point in decoding equals the header followed by the concatenation of all answer tokens produced so far. We do not reserve fixed answer slots inside the header; instead answers are accumulated at the end of the sequence as they are generated.

To track the role and origin of each token and make masking causal, we compute a small metadata tuple for each token index $i$ in $s$: a virtual position ID $pos^{\textrm{virt}}(i)$, a context identifier $ctx(i)\in\{0,1,\dots,J\}$ (with $ctx(i)=0$ indicating tokens in $p$), and a question identifier $qtn(i)\in\{0,1,\dots\, M\}$ (with $qtn(i)=0$ for tokens in $p$ and $c$). 

We distinguish between a \emph{virtual position} that encodes the per-triplet autoregressive ordering and the \emph{absolute position} that reflects the token's index in the concatenated sequence $s$. The virtual position of the $t$-th token of the answer for question $(j,k)$ is defined as the position that token would occupy if the triplet $(p,c_j,x_{j,k})$ were decoded in isolation: 
\begin{smallalign}
    pos^{\mathrm{virt}}\big(y^{j,k}_t\big) \;=\; |p| \;+\; |c_j| \;+\; |x_{j,k}| \;+\; (t-1),
\end{smallalign}
where  $|\cdot|$ denotes the token length of a sequence.
For header tokens inside $p$, $c_j$, or $x_{j,k}$, their virtual positions $pos^{\mathrm{virt}}(\cdot)$ cover these ranges:
\begin{smallalign}
    pos^{\mathrm{virt}}(p) &= [0\,,\, |p|), \\
    pos^{\mathrm{virt}}(c_j) &= [|p|\,,\, |p| + |c_j|), \\
    pos^{\mathrm{virt}}(x_{j,k}) &= [|p| + |c_j|\,,\, |p| + |c_j| + |x_{j,k}|).
\end{smallalign}
For answer tokens, $pos^{\mathrm{virt}}(\cdot)$ gives the local offset inside their parent triplet. Appendix \ref{app:abs_pos_id} provides a derivation for the absolute position IDs.

We define the masking logical expressions to use virtual positions for the causality constraint. Let $q,k$ be token indices in $s$ and $ctx(\cdot)$, $qtn(\cdot)$ be the context and question identifiers as previously defined. The elementary logical expressions become
\begin{smallalign}
    \mathrm{M_{caus.}}(q,k) &\;=\; \big(pos^{\mathrm{virt}}(q) \ge pos^{\mathrm{virt}}(k)\big), \label{eq:causalMask} \\ 
    \mathrm{M_{ctx}}(q,k) &\;=\; \big(ctx(q)=ctx(k)\big) \;\lor\; \big(ctx(k)=0\big), \\
    \mathrm{M_{qtn}}(q,k) &\;=\; \big(qtn(q)=qtn(k)\big) \;\lor\; \big(qtn(k)=0\big).
\end{smallalign}
The final mask logic is the conjunction
\begin{smallalign}
    \mathrm{M_{fin.}}(q,k) \;=\; \mathrm{M_{caus.}}(q,k) \;\land\; \mathrm{M_{ctx}}(q,k) \;\land\; \mathrm{M_{qtn}}(q,k)
\end{smallalign}
of the causal, context and question masks. The transformer's attention matrix mask is then set by
\begin{smallalign}
    M_{q,k} \;=\; \begin{cases} 0, & \text{if }\mathrm{M_{fin.}}(q,k)\text{ is true}, \\[4pt] -\infty, & \text{otherwise.} \end{cases}
\end{smallalign}
Negative infinity inputs result in zero activation post-softmax in the attention block. 

Because answers are appended at the end of the sequence and are labeled with their originating question and context, this mask guarantees that any answer token $y^{j,k}_t$ may attend only to the global instruction $p$ (via $ctx(k)=0$), to tokens in its parent context $c_j$ (via $ctx(q)=ctx(k)$), and to tokens in its own question/answer pair $x_{j,k}$ and previously generated answer tokens (via $qtn(q)=qtn(k)$). Tokens belonging to other questions or contexts fail at least one mask and are therefore blocked, preventing cross-question information leakage despite the single forward-pass execution.

\subsection{Parallel Decoding}
\label{sec:parallel-decoding}

Building on the hierarchical position and masking scheme described above, we now argue that the resulting forward pass is equivalent to independent autoregressive decoding over each triplet $(p,c_j,x_{j,k})$. The key point is that the attention mechanism for any output token $y^{j,k}_t$ is restricted by construction to exactly the same set of tokens it would see in the autoregressive case: the shared instruction $p$, the corresponding context $c_j$, the question $x_{j,k}$, and its own previously generated tokens $y^{j,k}_{<t}$. Tokens from other questions or contexts are masked and therefore have no influence.

Formally, let $q$ be the query vector for token $y^{j,k}_t$ and let $K$ be the matrix of key vectors over all tokens. Let $S$ denote the set of key indices that the autoregressive computation treats as valid, which the causal mask determines through $S = \{\, k \mid M_{\mathrm{CAUSAL}}[k] = 0 \,\}$. In the autoregressive setting, we have
\begin{smallalign}
    attn_{\mathrm{AR}}(y^{j,k}_t)
    &\;=\; \mathrm{softmax}\!\Big(\frac{qK^\top}{\sqrt{d_k}} + M_{\mathrm{CAUSAL}}\Big)V.
\end{smallalign}
The additive mask assigns $-\infty$ to every key outside $S$, which removes those keys from the softmax support without any explicit restriction of the softmax.
Under IPPD, the same attention score is computed over all keys but with a mask $M_{\mathrm{IPPD}}$ applied:
\begin{smallalign}
    attn_{\mathrm{IPPD}}(y^{j,k}_t)
    &\;=\; \mathrm{softmax}\!\Big(\frac{qK^\top}{\sqrt{d_k}} + M_{\mathrm{IPPD}}\Big)V,
\end{smallalign}
\begin{smallalign}
    M_{\mathrm{IPPD}}[k] \;=\; \begin{cases} 0, & k\in S,\\ -\infty, & k\notin S. \end{cases}
\end{smallalign}
Since adding $-\infty$ to a logit removes it from the softmax support, the resulting distribution is identical to that of the autoregressive case:
\begin{smallalign}
    attn_{\mathrm{IPPD}}(y^{j,k}_t) \;=\; attn_{\mathrm{AR}}(y^{j,k}_t).
\end{smallalign}
Thus, each token in the stacked prompt follows exactly the same computational path as in the independent autoregressive runs, establishing the equivalence of the two procedures.

Each forward pass at step $t$ generates the $t$-th token of each answer, which is appended to the prompt with the correct virtual position ID for the next pass. We stop appending tokens for answers that generate the end-of-sequence token, continuing the others until all answers are complete. This process is demonstrated in Fig.~\ref{fig:figure2}. Standard LLM frameworks such as Hugging Face Transformers \citep{wolf2020huggingfacestransformersstateoftheartnatural} natively support the multi-token output and custom attention masks required by IPPD. Consequently, IPPD requires no modifications to the model implementation, attention kernel and KV cache structure. This ensures broad compatibility and a significantly lower barrier to adoption compared to tree-based alternatives.

\subsection{Computational Efficiency} \label{sec:comp_efficiency}
IPPD allows for more efficient computation in the attention layers. We provide a high-level explanation in this section, with a more detailed analysis and comparison to Cascade inference available in \Cref{app:comp_efficiency_details}. At each attention layer, the attention computation involves multiplying $q$ and $K^T$, followed by a multiplication with $V$ post-softmax. These large multiplications incur a time cost for each byte of memory accessed to load these matrices, as well as a time cost for each floating point operation (FLOP) required for calculating the output. In autoregressive decoding, attention is almost exclusively bottlenecked by memory accesses, even for small models and large batch sizes \citep{Recasens2025MindTheGap}. This means that modern GPUs are underutilized because the ratio of FLOPs to bytes accessed from memory, known as \textit{arithmetic intensity}, is too low. IPPD alleviates this bottleneck by increasing arithmetic intensity.

Let $K_{p},K_{c_j},V_{p},V_{c_j}$ denote the keys and values belonging to the instruction tokens $p$ and the common context tokens $c_j$ respectively. These keys and values can be computed once and cached for all prompts in the batch, in a process known as prefix caching. Nevertheless, for each attention layer, batched autoregressive inference still requires each key and value to be accessed for every output token, as each prompt requires an independent attention calculation. Since IPPD combines all $\{x_{j,k}\}_{k=1}^{M_j}$ questions into one stacked prompt, it exposes reuse of $K_{p},K_{c_j},V_{p},V_{c_j}$ within a shared attention operation that generates $M_j$ output tokens. As the instruction $p$ and context $c_j$ usually contain the most tokens, this reuse can substantially reduce repeated memory traffic per output token. Although IPPD processes more FLOPs per output token, this tradeoff improves decoding throughput when attention is memory-bound. \Cref{app:comp_efficiency_details} quantifies this tradeoff and contrasts it with tree-based decoding strategies. We summarize the analysis into three key observations:

    (i) For tasks requiring a \textbf{single-token answer}, like multiple-choice QA (RACE in Table~\ref{tab:dataset_info}), IPPD is most effective with shorter contexts due to lower arithmetic intensity. 
    
    (ii) For tasks requiring a \textbf{multi-token answer} (NarrativeQA in Table~\ref{tab:dataset_info}), IPPD is effective for both short and long contexts, as arithmetic intensity is low regardless of context length.
    
    (iii) In CCQA scenarios characterized by \textbf{long shared contexts} and \textbf{short unique suffixes}, IPPD is better suited than tree-based decoding, as the overhead of combining partial attention scores in tree methods outweighs the few FLOPs IPPD consumes on masked attention entries.

\begin{table*}[ht!]
    \centering \small
    \begin{tabular}{@{}lcc|cccc@{}}
    \toprule
    & & & \multicolumn{4}{c}{Average token length}  \\
    \multicolumn{1}{c|}{Dataset} & \makecell{Number of \\ contexts} & \makecell{Avg. \\ Parallelism} & Instruction & Context & Question & Answer   \\ \midrule
    \multicolumn{1}{l|}{NarrativeQA (5s)}   & 355 & 29.73 & 2,754.00  & 737.45 & 11.97 & 5.81  \\
    \multicolumn{1}{l|}{SQuAD 2.0}   & 1,365 & 8.70 & 47.00   & 177.94 & 12.58 & 3.94   \\
    \multicolumn{1}{l|}{RACE}          & 1,407 & 3.51 & 40.00    & 344.30 & 39.81 & 1.00   \\
    \multicolumn{1}{l|}{LongHealth}    & 20 & 20.00 & 73.00     & 11,719.75 & 72.50 & 1.00  \\
    \bottomrule
    \end{tabular}
    \caption{Dataset statistics with the average token length of each prompt sub-section. (5s) means the instruction $p$ includes 5 few-shot examples. Average parallelism is the average number of questions sharing a common context, i.e., $\bar{M} = \frac{1}{J}\sum_{j=1}^{J} M_j$ for a dataset with $J$ contexts and $M_j$ questions per context $c_j$.}
    \label{tab:dataset_info}
\end{table*}

\subsection{Batch Inference Compatibility}
IPPD is fully compatible with batched inference. We can batch $b$ stacked prompts each with $\ell$ contexts. This allows us to trade off having shorter stacked prompts but lower intra-prompt parallelism. When using batched inference with IPPD, special care must be taken to maintain a uniform input length across stacked prompts. During prefill, the initial prompt is left-padded, as is normally done with standard batching. During decoding, each prompt may produce a variable number of output tokens per inference step. This is because (a) the number of questions or contexts may vary across stacked prompts, and (b) some answers may complete at an earlier step than others, thus reducing the number of outputs in the next step non-uniformly. To address this, we append pad tokens at each decoding step to match the input length of the prompt with the most answers currently being generated. We assign these pad tokens a virtual position ID of infinity (implemented as the maximum integer value), so that the causal mask in \Cref{eq:causalMask} prevents any attention. The virtual position IDs of other tokens also ignore the existence of pad tokens, so the batched output of IPPD is unchanged. We implement prefix caching within IPPD only for the dataset-wide instruction $p$. This affects NarrativeQA, where $p$ contains five few-shot examples, and allows its KV cache to be reused across batched stacked prompts. IPPD also covers workloads in which only some questions share a context. Questions with distinct contexts can be stacked in the same prompt, each carrying its own context, which still lowers the number of inference steps even though the gain is smaller than for questions that share a context. \Cref{tab:hyperparameters} shows that stacking up to six contexts per prompt is throughput-optimal on multi-token answer tasks such as SQuAD 2.0. Because every IPPD modification acts at the level of a single prompt, the method could compose with strategies such as length-based grouping or continuous batching. Supporting these strategies in a production serving system nevertheless requires backend-specific integration work.

\section{Experimental Setup}
\label{sec:exp_setup}
\paragraph{Datasets}

We evaluate IPPD on four CCQA datasets (Table \ref{tab:dataset_info}). First, we evaluate on NarrativeQA \citep{kocisky2018_narrativeqa}, a challenging dataset where answers are not required to be spans of the input. %
We employ 5-shot prompting \citep{liang2023holistic} to investigate performance on long prefixes. We select SQuAD 2.0 \citep{rajpurkar2018_squadv2} as an extractive QA benchmark with shorter contexts. We also evaluate on the multiple-choice benchmarks RACE \citep{lai2017_race} and LongHealth \citep{adams2025_longhealth} to cover a wide range of prefix and answer lengths combinations as well as domains. Although IPPD is best suited for large-scale industrial workloads, the scarcity of public datasets of such magnitude necessitates the use of these benchmarks as proxies. \Cref{tab:dataset_info} provides the important dataset statistics, while \Cref{app:datasets} contains additional details.

\begin{table*}[t!]
\centering\small
\aboverulesep=0ex
\belowrulesep=0.25ex

\begin{tabular}{@{}lcccccccc@{}}
\toprule
\multirow{3}{*}{Model}                                & \multicolumn{2}{c}{\multirow{2}{*}{NarrativeQA (5 shot)}} & \multicolumn{2}{c}{\multirow{2}{*}{SQuAD 2.0}} & \multicolumn{2}{c}{\multirow{2}{*}{RACE}} & \multicolumn{2}{c}{\multirow{2}{*}{LongHealth}}                                       \\
                                                      & \multicolumn{2}{c}{}                         & \multicolumn{2}{c}{}                         & \multicolumn{2}{c}{}                         & \multicolumn{2}{c}{}                         \\
                                                      & ROUGE-L    & EM\%                       & F1       & EM\%                        & Acc.\%         & EM\%                        & Acc.\%       & EM\%        \\
                            \midrule

\multirow{1}{*}{Qwen3-32B}  & $0.771_{+0.000}$      & \multicolumn{1}{c|}{96.5}      &   $0.649_{\textcolor{red}{-0.004}}$     & \multicolumn{1}{c|}{96.9}      &   $91.1_{+0.0}$    &   \multicolumn{1}{c|}{99.7}  &   $87.0_{\textcolor{red}{-0.5}}$    & \multicolumn{1}{c}{99.5}             \\

\multirow{1}{*}{OLMo-2-32B} & $0.752_{+0.000}$      & \multicolumn{1}{c|}{97.6}      &   $0.714_{+0.000}$     & \multicolumn{1}{c|}{98.1}      &   $90.6_{+0.0}$    &     \multicolumn{1}{c|}{99.8}  & -- & \multicolumn{1}{c}{--}            \\

\multirow{1}{*}{Phi4-14B}  & $0.702_{\textcolor{red}{-0.004}}$      & \multicolumn{1}{c|}{91.3}      &   $0.466_{\textcolor{red}{-0.005}}$    & \multicolumn{1}{c|}{95.8}      &   $88.0_{+0.0}$    &    \multicolumn{1}{c|}{99.9} &   $86.5_{\textcolor{Green}{+0.5}}$    & \multicolumn{1}{c}{99.3}              \\

\multirow{1}{*}{Qwen3-8B}  & $0.727_{+0.000}$      & \multicolumn{1}{c|}{97.5}      &   $0.598_{\textcolor{red}{-0.001}}$    & \multicolumn{1}{c|}{97.5}      &   $88.3_{\textcolor{red}{-0.1}}$    &    \multicolumn{1}{c|}{99.6}   &   $83.3_{+0.0}$    & \multicolumn{1}{c}{100}            \\

\multirow{1}{*}{Qwen3-4B-Instr.}  & $0.735_{+0.000}$      & \multicolumn{1}{c|}{95.7}      &   $0.659_{\textcolor{Green}{+0.002}}$    & \multicolumn{1}{c|}{98.1}      &   $87.4_{+0.0}$     &      \multicolumn{1}{c|}{99.7}   &   $83.0_{\textcolor{Green}{+0.5}}$    & \multicolumn{1}{c}{99.5}          \\

\multirow{1}{*}{Qwen3-1.7B}  & $0.679_{+0.000}$      & \multicolumn{1}{c|}{95.7}      &   $0.590_{\textcolor{Green}{+0.002}}$    & \multicolumn{1}{c|}{95.5}      &   $76.7_{\textcolor{red}{-0.1}}$    &       \multicolumn{1}{c|}{99.1}  &   $69.0_{\textcolor{red}{-0.3}}$    & \multicolumn{1}{c}{99.3}          \\
\bottomrule
                           
\end{tabular}
\caption{Quality metrics for standard batched inference, with the delta to IPPD outputs shown in subscript. EM (Exact Match)\% measures the percentage of batched inference and IPPD answers matching each other exactly. IPPD provides the same computational path for each token but numerical instabilities result in a near 100\% EM rate.
}
\label{tab:quality}
\end{table*}

\begin{figure*}[t!]
\centering
\small
\begin{subfigure}{0.48\textwidth}
\centering
\begin{tikzpicture}[scale=0.8]
\begin{axis}[
    ybar,
    bar width=5pt,
    bar shift=0pt,
    width=1.3\textwidth,
    height=4cm,
    ylabel={Speedup},
    ymin=0,
    ymax=35,
    xtick={1,2,3,4,5,6},
    xticklabels={Qwen3-1.7B, Qwen3-4B-I, Qwen3-8B, Phi4-14B, Qwen3-32B, OLMo2-32B},
    x tick label style={font=\small, rotate=30, anchor=east,xshift=0.7cm, yshift=-12pt},
    ylabel style={font=\footnotesize},
    title={NarrativeQA (5s)},
    title style={font=\normalsize\bfseries},
    grid=major,
    grid style={dashed, gray!20},
    enlarge x limits=0.12,
]

\addplot[fill=blue!90, draw=blue!100, line width=0.5pt] coordinates {
    (0.79, 1) (1.79, 1) (2.79, 1) (3.79, 1) (4.79, 1) (5.79, 1)
};

\addplot[fill=blue!50, draw=blue!65, line width=0.5pt] coordinates {
    (1, 17.22) (2, 19.92) (3, 15.08) (4, 15.85) (5, 14.71) (6, 13.24)
};

\addplot[fill=blue!20, draw=blue!30, line width=0.5pt] coordinates {
    (1.21, 23.33) (2.21, 28.67) (3.21, 33.01) (4.21, 28.71) (5.21, 22.38) (6.21, 26.91)
};

\end{axis}
\end{tikzpicture}
\end{subfigure}
\hfill
\begin{subfigure}{0.48\textwidth}
\centering
\begin{tikzpicture}[scale=0.8]
\begin{axis}[
    ybar,
    bar width=5pt,
    bar shift=0pt,
    width=1.3\textwidth,
    height=4cm,
    ymin=0,
    ymax=8,
    xtick={1,2,3,4,5,6},
    xticklabels={Qwen3-1.7B, Qwen3-4B-I, Qwen3-8B, Phi4-14B, Qwen3-32B, OLMo2-32B},
    x tick label style={font=\small, rotate=30, anchor=east,xshift=0.7cm, yshift=-12pt},
    ylabel style={font=\footnotesize},
    title={SQuAD 2.0},
    title style={font=\normalsize\bfseries},
    grid=major,
    grid style={dashed, gray!20},
    enlarge x limits=0.12,
]

\addplot[fill=teal!90, draw=teal!100, line width=0.5pt] coordinates {
    (0.79, 1) (1.79, 1) (2.79, 1) (3.79, 1) (4.79, 1) (5.79, 1)
};

\addplot[fill=teal!50, draw=teal!65, line width=0.5pt] coordinates {
    (1, 4.87) (2, 4.42) (3, 3.52) (4, 5.06) (5, 4.16) (6, 3.61)
};

\addplot[fill=teal!20, draw=teal!30, line width=0.5pt] coordinates {
    (1.21, 4.10) (2.21, 4.60) (3.21, 4.35) (4.21, 7.19) (5.21, 5.50) (6.21, 5.60)
};

\end{axis}
\end{tikzpicture}
\end{subfigure}

\begin{subfigure}{0.48\textwidth}
\centering
\begin{tikzpicture}[scale=0.8]
\begin{axis}[
    ybar,
    bar width=5pt,
    bar shift=0pt,
    width=1.3\textwidth,
    height=4cm,
    ylabel={Speedup},
    ymin=0,
    ymax=3,
    xtick={1,2,3,4,5,6},
    xticklabels={Qwen3-1.7B, Qwen3-4B-I, Qwen3-8B, Phi4-14B, Qwen3-32B, OLMo2-32B},
    x tick label style={font=\small, rotate=30, anchor=east,xshift=0.7cm, yshift=-12pt},
    ylabel style={font=\footnotesize},
    title={RACE},
    title style={font=\normalsize\bfseries},
    grid=major,
    grid style={dashed, gray!20},
    enlarge x limits=0.12,
]

\addplot[fill=brown!90, draw=brown!100, line width=0.5pt] coordinates {
    (0.79, 1) (1.79, 1) (2.79, 1) (3.79, 1) (4.79, 1) (5.79, 1)
};

\addplot[fill=brown!50, draw=brown!65, line width=0.5pt] coordinates {
    (1, 1.22) (2, 1.16) (3, 1.19) (4, 1.28) (5, 1.29) (6, 1.29)
};

\addplot[fill=brown!30, draw=brown!30, line width=0.5pt] coordinates {
    (1.21, 2.18) (2.21, 2.39) (3.21, 2.46) (4.21, 2.63) (5.21, 2.61) (6.21, 2.72)
};

\end{axis}
\end{tikzpicture}
\end{subfigure}
\hfill
\begin{subfigure}{0.48\textwidth}
\centering
\begin{tikzpicture}[scale=0.8]
\begin{axis}[
    ybar,
    bar width=5pt,
    bar shift=0pt,
    width=1.3\textwidth,
    height=4cm,
    ymin=0,
    ymax=18,
    xtick={1,2,3,4,5,6},
    xticklabels={Qwen3-1.7B, Qwen3-4B-I, Qwen3-8B, Phi4-14B, Qwen3-32B, OLMo2-32B},
    x tick label style={font=\small, rotate=30, anchor=east,xshift=0.7cm, yshift=-12pt},
    ylabel style={font=\footnotesize},
    title={LongHealth},
    title style={font=\normalsize\bfseries},
    grid=major,
    grid style={dashed, gray!20},
    enlarge x limits=0.12,
]

\addplot[fill=purple!90, draw=purple!100, line width=0.5pt] coordinates {
    (0.83, 1) (1.83, 1) (2.83, 1) (3.83, 1) (4.83, 1)
};

\addplot[fill=purple!50, draw=purple!65, line width=0.5pt] coordinates {
    (1, 13.89) (2, 15.69) (3, 16.10) (4, 15.95) (5, 15.67)
};

\addplot[fill=purple!20, draw=purple!30, line width=0.5pt] coordinates {
    (1.17, 8.48) (2.17, 8.75) (3.17, 9.79) (4.17, 11.48) (5.17, 10.72)
};

\end{axis}
\end{tikzpicture}
\end{subfigure}

\centering\scriptsize
\begin{tikzpicture}
\node[anchor=center] at (0,0) {
    \begin{tabular}{c}
    \colorbox{blue!90}{\textcolor{blue!90}{\rule{1.5pt}{1.5pt}}}\colorbox{teal!90}{\textcolor{teal!90}{\rule{1.5pt}{1.5pt}}}\colorbox{brown!90}{\textcolor{brown!90}{\rule{1.5pt}{1.5pt}}}\colorbox{purple!90}{\textcolor{purple!90}{\rule{1.5pt}{1.5pt}}} Batched Inference \quad
    \colorbox{blue!50}{\textcolor{blue!50}{\rule{1.5pt}{1.5pt}}}\colorbox{teal!50}{\textcolor{teal!50}{\rule{1.5pt}{1.5pt}}}\colorbox{brown!50}{\textcolor{brown!50}{\rule{1.5pt}{1.5pt}}}\colorbox{purple!50}{\textcolor{purple!50}{\rule{1.5pt}{1.5pt}}} Prefix Caching + PagedAttention \quad
    \colorbox{blue!20}{\textcolor{blue!20}{\rule{1.5pt}{1.5pt}}}\colorbox{teal!20}{\textcolor{teal!20}{\rule{1.5pt}{1.5pt}}}\colorbox{brown!20}{\textcolor{brown!20}{\rule{1.5pt}{1.5pt}}}\colorbox{purple!20}{\textcolor{purple!20}{\rule{1.5pt}{1.5pt}}} IPPD
    \end{tabular}
};
\end{tikzpicture}

\caption{Throughput measured in questions answered per second. Normalized throughput is relative to the standard batched inference baseline. IPPD outperforms prefix caching + PagedAttention for both short answer QA tasks (top), and for one of the two multiple-choice tasks (bottom). Darker colored shades represent batched inference (baseline), medium shades represent prefix caching + PagedAttention, and lighter shades represent IPPD.}
\label{fig:results}
\end{figure*}
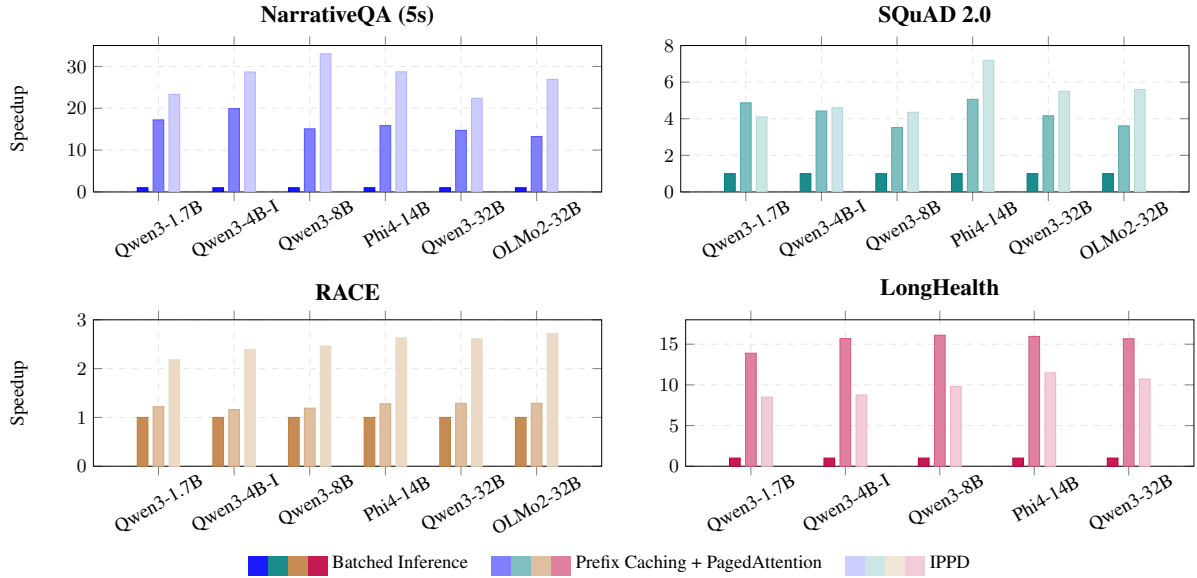

\paragraph{Models}
We evaluate open-source LLMs of varying sizes, namely, Qwen-3 (32B, 8B, 4B-Instruct-2507, 1.7B) \citep{qwen3technicalreport} using non-thinking mode for computational efficiency. We also evaluate Phi-4 14B \citep{phi4technicalreport} and OLMo-2-0325-32B-Instruct \citep{olmo2025_olmo2}. %
Models larger than 8B parameters are quantized to 4 bits to fit on a single GPU.

\paragraph{Baselines}
As the current version of IPPD is implemented using HuggingFace Transformers \citep{wolf2020huggingfacestransformersstateoftheartnatural}, we use Transformers' standard batched inference as our primary baseline. We also compare our method to inference using prefix caching and PagedAttention \citep{kwon2023efficient} (PC+PA), using vLLM as a backend. Like IPPD, these methods prevent KV recomputation and efficiently shares KV cache memory across questions with a shared context. Since vLLM is a production-oriented backend, it provides a number of \textit{orthogonal} optimizations not available with Transformers, such as advanced scheduling and custom efficient model code. As our goal is to directly compare IPPD against PC+PA specifically, we adjust the vLLM inference hyperparameters to most closely match our baseline system, with other confounding factors mitigated or removed. \Cref{app:inference_details} provides full experimental details, including hyperparameter selection for each method.  

\paragraph{Profiling and controlled ablations}
In addition to end-to-end throughput, we profile the prefill and decode phases separately to identify where IPPD reduces inference time. We also vary the number of questions sharing each context from 1 to 32 and increase context length from $1\times$ to $8\times$ to characterize how the available shared-context parallelism and prefill cost affect speedup. These experiments use controlled variants of the real benchmark examples rather than fully synthetic inputs. Their construction and complete results are described in \Cref{app:prefill_decode,app:controlled_ablations}.

\section{Results}

\Cref{tab:quality} shows that IPPD maintains near-identical performance to batched inference, with task-metric differences within one percentage point across all models and benchmarks. Exact match rates range from 91.3\% to 100\%, with 22 of 23 model--dataset combinations exceeding 95\%. \Cref{app:numerical_stability} explains the finite-precision arithmetic differences.

\Cref{fig:results} showcases the throughput of IPPD and prefix caching + PagedAttention (PC+PA) relative to standard batched inference across datasets and models. Absolute throughput numbers are available in Appendix \ref{app:add_results}. For the short answer generation datasets, IPPD consistently outperforms both batched inference and PC+PA. NarrativeQA throughput increases by up to 32X with Qwen3-8B. Because IPPD prefix-caches the dataset-wide instruction $p$ for NarrativeQA, which contains five few-shot examples, part of this speedup is attributable to reusing that prefix. IPPD nevertheless outperforms PC+PA by 2.2X despite PC+PA's more advanced cache management through the vLLM backend. IPPD also performs strongly with SQuAD 2.0, which features a much shorter instruction and average context length than NarrativeQA. IPPD's relative throughput is consistent across models, ranging from 4.1-5.6X for all models except Phi4-14B, where we measure a 7.2X relative throughput. Although IPPD performs well across the entire model range, it is strongest on larger models, as shown by throughput on SQuAD 2.0 being slightly lower than PC+PA with Qwen3-1.7B but convincingly higher for models with 8 billion parameters or more.This trend suggests that IPPD becomes increasingly valuable as attention computations scale.

For our multiple-choice datasets, IPPD consistently outperforms standard batched inference. RACE contains the lowest average number of questions per context, which results in a 2.1-2.7X relative throughput increase for IPPD. However, PC+PA fails to provide a substantial throughput increase, likely due to the short average context length limiting the effects of KV caching and limiting arithmetic intensity. By stacking six contexts per prompt, IPPD provides a 5.8X greater increase in throughput (+171\% vs. +29\%) with OLMo2-32B. This result validates our theoretical analysis in \Cref{app:comp_efficiency_details}, which states that attention can still be memory bottlenecked during prefill when contexts are short. In contrast, LongHealth features extremely long contexts, which have higher arithmetic intensity during prefill. IPPD is not as effective in this setting as PC+PA, as the advantages of efficient KV cache management for long prefixes outweigh the memory access reductions of IPPD when prefill is already arithmetically intense.

The phase-level profiling in \Cref{app:prefill_decode} confirms that IPPD reduces both prefill and decode time relative to batched autoregressive inference in every supported multi-token model--dataset setting. Decode-time reductions generally track the reduction in decoding steps, consistent with decode remaining memory-bandwidth bound despite the additional FLOPs in each IPPD step.

The question-count ablation in \Cref{app:question_ablation} identifies shared-context parallelism as the primary driver of IPPD's benefit: its win rate against PC+PA rises from 0\% at one question per context to 73\% at 16 and 32 questions, with LongHealth accounting for five of the six remaining losses. Although longer contexts reduce IPPD's relative advantage, the context-length ablation in \Cref{app:context_ablation} shows that it still wins 75\% of supported comparisons at $8\times$ context length.

\section{When to Use IPPD}
\label{sec:when_to_use}
IPPD performs best in throughput-oriented workloads where many independently answerable questions are available together. The number of questions per context is the clearest practical indicator of its benefit, as examined in \Cref{app:question_ablation}. For multi-token outputs, IPPD remains effective across context lengths because decoding is memory-bandwidth bound. Its advantage narrows when answers are long relative to the shared context and question-specific suffixes dominate the stacked prompt, as analyzed in \Cref{app:prefill_decode,app:context_ablation,app:comp_efficiency_details}. For single-token outputs, short contexts such as RACE favor IPPD, whereas long contexts have higher prefill arithmetic intensity and can favor PC+PA, as on LongHealth in \Cref{fig:results}. \Cref{tab:decision} summarizes these regimes. With short contexts and long generations, neither method targets the dominant question-specific decoding work, so standard batched inference or a specialized method such as speculative decoding \citep{leviathan2023_spec_decoding} may be preferable. A large class of practically important CCQA workloads fall into the regime preferred by IPPD, such as extracting many distinct fields from unstructured documents, product and web data, or validating documents against a set of requirements for quality assurance. IPPD can also apply inside workloads that appear sequential, since reasoning problems can often be decomposed into sub-questions answered before any cross-dependency arises.

\begin{table}[t]
\centering\small
\begin{tabular}{@{}lcc@{}}
\toprule
 & Short context & Long context \\ \midrule
Single-token answer & \textbf{IPPD} & PC+PA \\
Multi-token answer & \textbf{IPPD} & \textbf{IPPD} \\
Long generation & Standard inf. & PC+PA \\
\bottomrule
\end{tabular}
\caption{Recommended inference method by answer length and context length, assuming several questions share each context.}
\label{tab:decision}
\end{table}

\section{Conclusion}

We introduce IPPD, a novel approach that accelerates LLM inference for tasks with shared context such as CCQA. By manipulating position IDs and attention masks, IPPD can decode the next token for all questions in parallel without requiring any modifications to the LLM architecture. IPPD is fully compatible with batched inference, handling multiple prompts with different contexts simultaneously. IPPD achieves up to 7X speedup without sacrificing model performance. Experimental results demonstrate that IPPD outperforms batched autoregressive decoding on every benchmark dataset and model size we test, and outperforms prefix caching with PagedAttention in most evaluated settings. While our experiments focus on CCQA, the core concept of IPPD can extend to broader shared-context generation tasks such as recommendation systems and multi-aspect information extraction. Future work could explore combining IPPD with orthogonal efficiency techniques in systems such as vLLM to further reduce inference costs.

\clearpage
\section*{Limitations}
\label{sec:limitations}
Our evaluation targets offline workloads in which the full question set is available in advance, and IPPD requires that questions sharing a context be answerable independently, so tasks whose answers condition one another fall outside its scope. \Cref{sec:when_to_use} discusses a way to apply IPPD inside nominally sequential tasks, which we do not evaluate. IPPD's advantage also narrows in two regimes. Very long contexts paired with single-token answers raise arithmetic intensity during prefill, where prefix caching with PagedAttention outperforms IPPD on LongHealth. Generations that are long relative to the shared context reduce the advantage as well, as noted in \Cref{app:comp_efficiency_details}, though we do not measure this regime.

IPPD is implemented on top of HuggingFace Transformers for ease of experimentation and low-level control during inference. Unlike vLLM, this implementation is not optimized for high-throughput serving applications. As such, the reported inference speed likely understates IPPD’s attainable efficiency. We leave the integration of IPPD with vLLM as a promising opportunity for future work. Since IPPD modifies the attention mask, it is not directly compatible with FlashAttention \citep{dao2023_flashattention2}. However, it is compatible with the new implementation of FlexAttention \citep{dong2024flexattentionprogrammingmodel}, which provides similar benefits to FlashAttention with the flexibility of custom attention masks. IPPD is implemented with both FlexAttention and standard scaled dot-product attention (SDPA), since the latter performs better in most of our experiments. This may change in the future as support for FlexAttention matures and becomes better optimized. We do not evaluate every model size on all hardware platforms, so throughput may vary for untested model--hardware combinations.

We compare against standard batched inference and prefix caching with PagedAttention. Specialized shared-prefix and tree-based systems such as Cascade inference, Hydragen, DeFT and FastTree lack implementations compatible with the model families and inference platforms used here, so we do not compare against them. We have not implemented or evaluated IPPD with production batching policies, which remains future work.

\newpage

\bibliography{ref}

@inproceedings{Brinkmann2024_ExtractGPT,
author = {Brinkmann, Alexander and Shraga, Roee and Bizer, Christian},
title = {ExtractGPT: Exploring the Potential of Large Language Models for Product Attribute Value Extraction},
year = {2024},
publisher = {Springer-Verlag},
booktitle = {Information Integration and Web Intelligence: 26th International Conference, IiWAS 2024, Proceedings, Part I},
pages = {38--52},
numpages = {15},
}

@inproceedings{leviathan2023_spec_decoding,
title={Fast Inference from Transformers via Speculative Decoding}, 
author={Yaniv Leviathan and Matan Kalman and Yossi Matias},
year={2023},
booktitle = {Proceedings of the 40th International Conference on Machine Learning},
volume = 	 {202},
pages = 	 {19274--19286},
}

@inproceedings{Fu2024_lookahead_decoding,
author = {Fu, Yichao and Bailis, Peter and Stoica, Ion and Zhang, Hao},
title = {Break the sequential dependency of LLM inference using Lookahead Decoding},
year = {2024},
booktitle = {Proceedings of the 41st International Conference on Machine Learning},
volume = 	 {235},
pages = 	 {14060--14079},
}

@unpublished{liu2024_APAR,
      title={APAR: LLMs Can Do Auto-Parallel Auto-Regressive Decoding}, 
      author={Mingdao Liu and Aohan Zeng and Bowen Wang and Peng Zhang and Jie Tang and Yuxiao Dong},
      year={2024},
      note = {arXiv preprint: arXiv 2401.06761},    
}

@inproceedings{Cai2024_medusa,
author = {Cai, Tianle and Li, Yuhong and Geng, Zhengyang and Peng, Hongwu and Lee, Jason D. and Chen, Deming and Dao, Tri},
title = {MEDUSA: Simple LLM inference acceleration framework with multiple decoding heads},
year = {2024},
booktitle = {Proceedings of the 41st International Conference on Machine Learning},
volume = {235},
pages = {5209--5235},
}

@inproceedings{
ning2024_skeletonofthought,
title={Skeleton-of-Thought: Prompting {LLM}s for Efficient Parallel Generation},
author={Xuefei Ning and Zinan Lin and Zixuan Zhou and Zifu Wang and Huazhong Yang and Yu Wang},
booktitle={The Twelfth International Conference on Learning Representations},
year={2024},
}

@unpublished{qwen3technicalreport,
      title={Qwen3 Technical Report}, 
      author={An Yang and Anfeng Li and Baosong Yang and Beichen Zhang and Binyuan Hui and Bo Zheng and Bowen Yu and Chang Gao and Chengen Huang and Chenxu Lv and Chujie Zheng and Dayiheng Liu and Fan Zhou and Fei Huang and Feng Hu and Hao Ge and Haoran Wei and Huan Lin and Jialong Tang and Jian Yang and Jianhong Tu and Jianwei Zhang and Jianxin Yang and Jiaxi Yang and Jing Zhou and Jingren Zhou and Junyang Lin and Kai Dang and Keqin Bao and Kexin Yang and Le Yu and Lianghao Deng and Mei Li and Mingfeng Xue and Mingze Li and Pei Zhang and Peng Wang and Qin Zhu and Rui Men and Ruize Gao and Shixuan Liu and Shuang Luo and Tianhao Li and Tianyi Tang and Wenbiao Yin and Xingzhang Ren and Xinyu Wang and Xinyu Zhang and Xuancheng Ren and Yang Fan and Yang Su and Yichang Zhang and Yinger Zhang and Yu Wan and Yuqiong Liu and Zekun Wang and Zeyu Cui and Zhenru Zhang and Zhipeng Zhou and Zihan Qiu},
      year={2025},
      note = {arXiv preprint: arXiv 2505.09388}, 
}

@unpublished{phi4technicalreport,
      title={Phi-4 Technical Report}, 
      author = {Abdin, Marah I and Aneja, Jyoti and Behl, Harkirat and Bubeck, Sébastien and Eldan, Ronen and Gunasekar, Suriya and Harrison, Michael and Hewett, Russell J. and Javaheripi, Mojan and Kauffmann, Piero and Lee, James R. and Lee, Yin Tat and Li, Yuanzhi  and Liu, Weishung and Mendes, Caio CT and Nguyen, Anh and Price, Eric and de Rosa, Gustavo and Saarikivi, Olli and Salim, Adil and Shah, Shital and Wang, Xin and Ward, Rachel and Wu, Yue and Yu, Dingli and Zhang, Cyril and Zhang, Yi},
      year={2024},
      note = {arXiv preprint: arXiv 2412.08905}, 
}

@inproceedings{dao2023_flashattention2,
title={FlashAttention-2: Faster Attention with Better Parallelism and Work Partitioning},
author={Tri Dao},
booktitle={The Twelfth International Conference on Learning Representations},
year={2024},
}

@article{kocisky2018_narrativeqa,
    title = "The {N}arrative{QA} Reading Comprehension Challenge",
    author = "Ko{\v{c}}isk{\'y}, Tom{\'a}{\v{s}}  and
      Schwarz, Jonathan  and
      Blunsom, Phil  and
      Dyer, Chris  and
      Hermann, Karl Moritz  and
      Melis, G{\'a}bor  and
      Grefenstette, Edward",
    journal = "Transactions of the Association for Computational Linguistics",
    volume = "6",
    year = "2018",
    publisher = "MIT Press",
    pages = "317--328",
}

@inproceedings{lai2017_race,
    title = "{RACE}: Large-scale {R}e{A}ding Comprehension Dataset From Examinations",
    author = "Lai, Guokun  and
      Xie, Qizhe  and
      Liu, Hanxiao  and
      Yang, Yiming  and
      Hovy, Eduard",
    booktitle = "Proceedings of the 2017 Conference on Empirical Methods in Natural Language Processing",
    year = "2017",
    publisher = "Association for Computational Linguistics",
    pages = "785--794",
}

@inproceedings{rajpurkar2018_squadv2,
    title = "Know What You Don{'}t Know: Unanswerable Questions for {SQ}u{AD}",
    author = "Rajpurkar, Pranav  and
      Jia, Robin  and
      Liang, Percy",
    booktitle = "Proceedings of the 56th Annual Meeting of the Association for Computational Linguistics (Volume 2: Short Papers)",
    year = "2018",
    publisher = "Association for Computational Linguistics",
    pages = "784--789",
}

@inproceedings{yang-etal-2018-hotpotqa,
    title = "{H}otpot{QA}: A Dataset for Diverse, Explainable Multi-hop Question Answering",
    author = "Yang, Zhilin  and
      Qi, Peng  and
      Zhang, Saizheng  and
      Bengio, Yoshua  and
      Cohen, William  and
      Salakhutdinov, Ruslan  and
      Manning, Christopher D.",
    booktitle = "Proceedings of the 2018 Conference on Empirical Methods in Natural Language Processing",
    year = "2018",
    publisher = "Association for Computational Linguistics",
    pages = "2369--2380",
}

@article{adams2025_longhealth,
  title={Longhealth: A question answering benchmark with long clinical documents},
  author={Adams, Lisa and Busch, Felix and Han, Tianyu and Excoffier, Jean-Baptiste and Ortala, Matthieu and L{\"o}ser, Alexander and Aerts, Hugo JWL and Kather, Jakob Nikolas and Truhn, Daniel and Bressem, Keno},
  journal={Journal of Healthcare Informatics Research},
  volume={9},
  pages={280--296},
  year={2025},
  publisher={Springer}
}

@unpublished{olmo2025_olmo2,
      title={2 OLMo 2 Furious}, 
      author={Team OLMo and Pete Walsh and Luca Soldaini and Dirk Groeneveld and Kyle Lo and Shane Arora and Akshita Bhagia and Yuling Gu and Shengyi Huang and Matt Jordan and Nathan Lambert and Dustin Schwenk and Oyvind Tafjord and Taira Anderson and David Atkinson and Faeze Brahman and Christopher Clark and Pradeep Dasigi and Nouha Dziri and Allyson Ettinger and Michal Guerquin and David Heineman and Hamish Ivison and Pang Wei Koh and Jiacheng Liu and Saumya Malik and William Merrill and Lester James V. Miranda and Jacob Morrison and Tyler Murray and Crystal Nam and Jake Poznanski and Valentina Pyatkin and Aman Rangapur and Michael Schmitz and Sam Skjonsberg and David Wadden and Christopher Wilhelm and Michael Wilson and Luke Zettlemoyer and Ali Farhadi and Noah A. Smith and Hannaneh Hajishirzi},
      year={2025},
      note = {arXiv preprint: arXiv 2501.00656},  
}

@inproceedings{kwon2023efficient,
  title={Efficient Memory Management for Large Language Model Serving with PagedAttention},
  author={Woosuk Kwon and Zhuohan Li and Siyuan Zhuang and Ying Sheng and Lianmin Zheng and Cody Hao Yu and Joseph E. Gonzalez and Hao Zhang and Ion Stoica},
  booktitle={Proceedings of the ACM SIGOPS 29th Symposium on Operating Systems Principles},
  year={2023},
  pages = {611--626},
}

@inproceedings{ye-etal-2024-chunkattention,
    title = "{C}hunk{A}ttention: Efficient Self-Attention with Prefix-Aware {KV} Cache and Two-Phase Partition",
    author = "Ye, Lu  and
      Tao, Ze  and
      Huang, Yong  and
      Li, Yang",
    booktitle = "Proceedings of the 62nd Annual Meeting of the Association for Computational Linguistics (Volume 1: Long Papers)",
    year = "2024",
    publisher = "Association for Computational Linguistics",
    pages = "11608--11620",
}

@misc{cascade-inference,
    title = {Cascade Inference: Memory Bandwidth Efficient Shared Prefix Batch Decoding},
    url = {https://flashinfer.ai/2024/02/02/cascade-inference.html},
    author = {Ye, Zihao and Lai, Ruihang and Lu, Bo-Ru and Lin, Chien-Yu and Zheng, Size and Chen, Lequn and Chen, Tianqi and Ceze, Luis},
    year = {2024}
}

@inproceedings{Recasens2025MindTheGap,
  title={Mind the Memory Gap: Unveiling GPU Bottlenecks in Large-Batch LLM Inference},
  author={Pol G. Recasens and Ferran Agull{\'o} and Yue Zhu and Chen Wang and Eun Kyung Lee and Olivier Tardieu and Jordi Torres and Josep Ll. Berral},
  booktitle={2025 IEEE 18th International Conference on Cloud Computing (CLOUD)}, 
  year={2025},
  pages={277-287},
}

@inproceedings{dong2024flexattentionprogrammingmodel,
 author = {Dong, Juechu and FENG, BOYUAN and Guessous, Driss and Liang, Yanbo and He, Horace},
 booktitle = {Proceedings of Machine Learning and Systems},
 title = {FlexAttention: A Programming Model for Generating Fused Attention Variants.},
 volume = {7},
 year = {2025}
}

@article{zhang2025comprehensivesurveyprocessorientedautomatic,
      title={A Comprehensive Survey on Process-Oriented Automatic Text Summarization with Exploration of LLM-Based Methods}, 
      author={Yang Zhang and Hanlei Jin and Dan Meng and Jun Wang and Jinghua Tan},
      year = {2026}, 
      volume = {663},
      pages = {131928},
      journal = {Neurocomputing},
}

@article{yi2025surveyrecentadvancesllmbased,
      title={A Survey on Recent Advances in LLM-Based Multi-turn Dialogue Systems}, 
      author={Zihao Yi and Jiarui Ouyang and Zhe Xu and Yuwen Liu and Tianhao Liao and Haohao Luo and Ying Shen},
      year={2025},
      journal = {ACM Computing Surveys},
      volume = {58},
      articleno = {148},
}

@misc{yue2025surveylargelanguagemodel,
      title={A Survey of Large Language Model Agents for Question Answering}, 
      author={Murong Yue},
      year={2025},
      note = {arXiv preprint: arXiv 2503.19213}, 
}

@misc{wolf2020huggingfacestransformersstateoftheartnatural,
      title={HuggingFace's Transformers: State-of-the-art Natural Language Processing}, 
      author={Thomas Wolf and Lysandre Debut and Victor Sanh and Julien Chaumond and Clement Delangue and Anthony Moi and Pierric Cistac and Tim Rault and Rémi Louf and Morgan Funtowicz and Joe Davison and Sam Shleifer and Patrick von Platen and Clara Ma and Yacine Jernite and Julien Plu and Canwen Xu and Teven Le Scao and Sylvain Gugger and Mariama Drame and Quentin Lhoest and Alexander M. Rush},
      year={2020},
      note = {arXiv preprint: arXiv 1910.03771}, 
}

@article{
liang2023holistic,
title={Holistic Evaluation of Language Models},
author={Percy Liang and Rishi Bommasani and Tony Lee and Dimitris Tsipras and Dilara Soylu and Michihiro Yasunaga and Yian Zhang and Deepak Narayanan and Yuhuai Wu and Ananya Kumar and Benjamin Newman and Binhang Yuan and Bobby Yan and Ce Zhang and Christian Alexander Cosgrove and Christopher D Manning and Christopher Re and Diana Acosta-Navas and Drew Arad Hudson and Eric Zelikman and Esin Durmus and Faisal Ladhak and Frieda Rong and Hongyu Ren and Huaxiu Yao and Jue WANG and Keshav Santhanam and Laurel Orr and Lucia Zheng and Mert Yuksekgonul and Mirac Suzgun and Nathan Kim and Neel Guha and Niladri S. Chatterji and Omar Khattab and Peter Henderson and Qian Huang and Ryan Andrew Chi and Sang Michael Xie and Shibani Santurkar and Surya Ganguli and Tatsunori Hashimoto and Thomas Icard and Tianyi Zhang and Vishrav Chaudhary and William Wang and Xuechen Li and Yifan Mai and Yuhui Zhang and Yuta Koreeda},
journal={Transactions on Machine Learning Research},
issn={2835-8856},
year={2023},
}

@inproceedings{pan2025marconiprefixcachingera,
 author = {Pan, Rui and Wang, Zhuang and Jia, Zhen and Karakus, Can and Zancato, Luca and Dao, Tri and Wang, Yida and Netravali, Ravi},
 booktitle = {Proceedings of Machine Learning and Systems},
 title = {Marconi: Prefix Caching for the Era of Hybrid LLMs},
 volume = {7},
 year = {2025}
}

@inproceedings{
juravsky2024hydragen,
title={Hydragen: High-Throughput {LLM} Inference with Shared Prefixes},
author={Jordan Juravsky and Bradley Brown and Ryan Saul Ehrlich and Daniel Y Fu and Christopher Re and Azalia Mirhoseini},
booktitle={Workshop on Efficient Systems for Foundation Models II @ ICML},
year={2024},
}

@misc{IDPLeaderboard,
  title={IDPLeaderboard: A Unified Leaderboard for Intelligent Document Processing Tasks},
  author={Souvik Mandal and Nayancy Gupta and Ashish Talewar and Paras Ahuja and Prathamesh Juvatkar and Gourinath Banda},
  howpublished={https://idp-leaderboard.org},
  year={2025},
}

@inproceedings{
yao2025deft,
title={De{FT}: Decoding with Flash Tree-attention for Efficient Tree-structured {LLM} Inference},
author={Jinwei Yao and Kaiqi Chen and Kexun Zhang and Jiaxuan You and Binhang Yuan and Zeke Wang and Tao Lin},
booktitle={The Thirteenth International Conference on Learning Representations},
year={2025},
}

@inproceedings{pan2025fasttree,
 author = {Pan, Zaifeng and Ding, Yitong and Guan, Yue and Wang, Zheng and Yu, Zhongkai and Tang, Xulong and Wang, Yida and Ding, Yufei},
 booktitle = {Proceedings of Machine Learning and Systems},
 title = {FastTree: Optimizing Attention Kernel and Runtime for Tree-Structured LLM Inference},
 volume = {7},
 year = {2025}
}

@misc{wang2025flashforgeultraefficientprefixawareattention,
      title={FlashForge: Ultra-Efficient Prefix-Aware Attention for LLM Decoding}, 
      author={Zhibin Wang and Rui Ning and Chao Fang and Zhonghui Zhang and Xi Lin and Shaobo Ma and Mo Zhou and Xue Li and Zhongfeng Wang and Chengying Huan and Rong Gu and Kun Yang and Guihai Chen and Sheng Zhong and Chen Tian},
      year={2025},
      note = {arXiv preprint: arXiv 2505.17694},  
}

\clearpage
\appendix

\section*{\centering Appendix}

\section{Method Details}
\subsection{Position IDs and Attention Mask} \label{app:abs_pos_id}

In the stacked sequence $s$, each token has an \emph{absolute position} determined by its literal index in the concatenated prompt. As answers are appended sequentially, absolute positions grow continuously across all contexts and questions. Formally, the absolute position of the $t$-th token of the $k$-th question under context $j$ is
\begin{align}
    pos\big(y^{j,k}_t\big)
    ={}& L_{\mathrm{hdr}}
    + {\sum_{\substack{(j',k')\in\mathcal{Q}\\
    \mathrm{rank}(j',k')<\mathrm{rank}(j,k)}} T_{j',k'}} \notag\\
    &+ (t-1) + 1,
\end{align}
where $L_{\mathrm{hdr}}$ is the length of the instruction plus all context and question headers, $\mathcal{Q}=\big((1,1),(1,2),\dots,(J,M_J)\big)$ is the fixed ordering of question pairs in the prompt, and $T_{j',k'}$ is the token length of the generated answer for question $(j',k')$. This absolute index reflects the physical layout of the stacked sequence.

In contrast, our method introduces a \emph{virtual position} $pos^{\mathrm{virt}}(\cdot)$, which maps tokens back to positions that are local to their own instruction, context, and question triplet. For a generated token $y^{j,k}_t$, the virtual position is
\begin{align}
    pos^{\mathrm{virt}}\big(y^{j,k}_t\big)
    ={}& |p| + \mathrm{len}(c_j) \notag\\
    &+ |x_{j,k}| + (t-1).
\end{align}
Header tokens are similarly assigned virtual positions aligned with their logical block boundaries. 
By mapping absolute indices into logically local virtual positions, we ensure that attention remains causal within each sequence, even if two sequences are physically distant in $s$. In other words, $pos^{\mathrm{virt}}(\cdot)$ preserves the \textit{illusion} that each triplet $(p, c_j, x_{j,k})$ is decoded autoregressively in isolation, while still allowing parallel execution within a single stacked prompt.

\subsection{Computational Efficiency} \label{app:comp_efficiency_details}
We provide a deeper analysis of the FLOP and memory access trade-off of IPPD against prefix caching and PagedAttention (PC+PA). For this section, we assume that IPPD uses one context $c_j$, with $M_j$ questions $\{x_{j,k'}\}_{k'=1}^{M_j}$ stacked in one prompt. We calculate how FLOPs and memory accesses scale with the length of $p,c_j,$ and $x_{j,k}$, which we write as $|\cdot|$. We consider only one attention head of one attention layer, as the computation pattern for each head is the same. The FLOPs and bytes of memory accessed depend on the sequence length of the input and the hidden dimension of the query, key and value associated to each token. Since the hidden dimension is static, we focus on the sequence-length dependent dimension of $q,K$ and $V$, which we denote as $dim(\cdot)$. \Cref{tab:KV_size} contains $dim(q)$ and $dim(K,V)$ for both methods during different stages of inference. The memory accesses required for each attention computation are linearly proportional to the dimensions of $q,K$ and $V$, so we write that the bytes of memory accessed are $\mathcal{O}(dim(q) + dim(K,V))$, ignoring scaling constants for simplicity. Similarly, the FLOPs required for multiplication during attention are $\mathcal{O}(dim(q) \times dim(K,V))$. The arithmetic intensity is defined as the ratio of FLOPs to memory accesses, which scales as
\[
\mathcal{O}\left(\frac{dim(q) \times dim(K,V)}{dim(q) + dim(K,V)}\right).
\]
Low arithmetic intensity signifies that memory accesses are causing a performance bottleneck, which \citeauthor{Recasens2025MindTheGap} show is often the case with LLM inference. The following two sections compare the approximated FLOPs and memory accesses of PC+PA and IPPD for both prefill and decode. For a context $c_j$ and an inference step $t$, we consider the total FLOPs and memory accesses for computing attention for one output token of every question, $\{y^{j,k'}_{t}\}_{k'=1}^{M_j}$.

\subsubsection{Prefill}\label{app:prefill_efficiency} The first prefill with PC+PA for the prompt answering $x_{j,k}$ requires $dim(q)=dim(K,V)=|p|+|c_j|+|x_{j,k}|$, as the prefix cache has not yet been computed. In subsequent prefills, we only require $dim(q)=|x_{j,k}|$ since the instruction and context are in the prefix cache. The most significant FLOP cost is during the first prefill, which scales as $\mathcal{O}((|p|+|c_j|+|x_{j,k}|)^2)$. $M_j$ attention operations are required to prefill all questions, each requiring the full $K$ and $V$. Therefore, the memory accesses scale as $\mathcal{O}(M_j\times(|p|+|c_j|+|x_{j,k}|))$. \textbf{For IPPD}, the FLOP cost scales as $\mathcal{O}((|p|+|c_j|+\sum_{k'=1}^{M_j}|x_{j,k'}|)^2)$, since the prompt contains all questions. Under our operation-level model, memory traffic scales as $\mathcal{O}(|p|+|c_j|+\sum_{k'=1}^{M_j}|x_{j,k'}|)$ because the questions are processed through a shared attention operation. Writing $X_j=\sum_{k'=1}^{M_j}|x_{j,k'}|$, the tradeoff obtained by IPPD is a \textit{reduction} in memory accesses by a factor of $M_j$, in exchange for an \textit{increase} in FLOPs by a factor of
\[
\left(
\frac{|p|+|c_j|+X_j}
{|p|+|c_j|+|x_{j,k}|}
\right)^2.
\]
There are two scenarios where this is advantageous:
\begin{enumerate}
    \item When $|p|+|c_j| \gg \sum_{k'=1}^{M_j}|x_{j,k'}|$, and $M_j \gg 1$, the relative increase in FLOPs is minor compared to the reduction in memory accesses.
    \item Since arithmetic intensity (FLOPs / mem. access ratio) scales as $\mathcal{O}(|p|+|c_j|+|x_{j,k}|)$, shorter $p$, $c_j$, and $x_{k,j}$ result in lower arithmetic intensity. This scenario is more memory bottlenecked, so it benefits more from the reduction in memory accesses.
\end{enumerate}
We observe in \Cref{tab:dataset_info} that the RACE dataset matches both of these scenarios: the contexts are short, but the questions are even shorter. \Cref{fig:results} shows IPPD significantly outperforming PC+PA in this task. For the LongHealth dataset, the contexts are extremely large ($>10,000$ tokens on average), which increases arithmetic intensity linearly. This dataset is therefore more likely to be FLOP bottlenecked, and not benefit as much from IPPD. We observe in our results that PC+PA outperforms IPPD on this dataset. Both RACE and LongHealth require only one token per answer, so their performance characteristics are solely determined by the prefill stage.

\begin{table*}[t!]
\centering\small
\aboverulesep=0ex
\belowrulesep=0.25ex

\begin{tabular}{@{}llccc@{}}
\toprule
\multirow{3}{*}{Method}  &  & \multicolumn{2}{c}{\multirow{2}{*}{Prefill}} & \multirow{3}{*}{Decode}\\
 &  &     &    \\
 & & First  & Subsequent    &          \\  \midrule   
\multirow{6}{*}{PC+PA} & \multicolumn{1}{c|}{\multirow{2}{*}{$dim(q)$}} & \multirow{2}{*}{$|p|+|c_j|+|x_{j,k}|$} & \multicolumn{1}{c|}{\multirow{2}{*}{$|x_{j,k}|$}} & \multirow{2}{*}{$1$} \\ 
& \multicolumn{1}{c|}{}& & \multicolumn{1}{c|}{}\\
                        & \multicolumn{1}{c|}{\multirow{2}{*}{$dim(K,V)$}} &  \multicolumn{2}{c|}{\multirow{2}{*}{$|p|+|c_j|+|x_{j,k}|$}} & \multirow{2}{*}{\makecell[c]{$|p|+|c_j|+|x_{j,k}|$\\$+|y^{j,k}_{<t}|$}} \\ 
&\multicolumn{1}{c|}{} & & \multicolumn{1}{c|}{}\\ 
& \multicolumn{1}{c|}{\multirow{2}{*}{Frequency}}&  \multicolumn{2}{c|}{\multirow{2}{*}{Once per $x_{j,k}$}} &  \multirow{2}{*}{Once per $y^{j,k}_{t}$} \\
& \multicolumn{1}{c|}{}& & \multicolumn{1}{c|}{}\\ \midrule   
\multirow{6}{*}{IPPD} & \multicolumn{1}{c|}{\multirow{2}{*}{$dim(q)$}} & \multirow{2}{*}{$|p|+|c_j|+\sum_{k'=1}^{M_j}|x_{j,k'}|$} & \multicolumn{1}{c|}{\multirow{2}{*}{$|c_j|+\sum_{k'=1}^{M_j}|x_{j,k'}|$}} & \multirow{2}{*}{$M_j^*$} \\ 
& \multicolumn{1}{c|}{}& & \multicolumn{1}{c|}{}\\
                        & \multicolumn{1}{c|}{\multirow{2}{*}{$dim(K,V)$}} & \multicolumn{2}{c|}{\multirow{2}{*}{$|p|+|c_j|+\sum_{k'=1}^{M_j}|x_{j,k'}|$}}  & \multirow{2}{*}{\makecell[c]{$|p|+|c_j|$\\$+\sum_{k'=1}^{M_j}(|x_{j,k'}|+|y^{j,k'}_{<t}|)$}} \\ 
& \multicolumn{1}{c|}{} & & \multicolumn{1}{c|}{}\\  
& \multicolumn{1}{c|}{\multirow{2}{*}{Frequency}}&  \multicolumn{2}{c|}{\multirow{2}{*}{Once per $c_{j}$}} &  \multirow{2}{*}{Once per set $\{y^{j,k'}_{<t}\}_{k'=1}^{M_j^*}$} \\ 
& \multicolumn{1}{c|}{}& & \multicolumn{1}{c|}{}\\ \bottomrule       
\end{tabular}
\caption{Summary of the matrix dimensions for queries $q$ and keys/values $K$ and $V$ for a single attention head, using prefix caching + PagedAttention (PC+PA) and IPPD with one context per prompt. $dim(\cdot)$ denotes the dimension that varies with sequence length, ignoring the hidden dimension. Prefill denotes the first forward pass of each prompt, with the First column representing the calculation of the prefix cache. Subsequent prefills leverage this cache to avoid recomputations. The decode column represents the autoregressive decoding stage. The Frequency row indicates how often attention must be performed using these matrix dimensions. $M_j$ denotes the number of questions in context $c_j$. The $^*$ indicates that the value may decrease after multiple inference steps, as some answers complete generation before others. The table shows that IPPD's stacked prompts require larger dimensions but fewer attention calculations.}
\label{tab:KV_size}
\end{table*}

\subsubsection{Decode} During decode, PC+PA contains a single query, as tokens are generated autoregressively. However, that query must attend to all prior tokens, which requires a separate memory access to the keys and values of $p$ and $c_j$, $x_{j,k}$ and partial output $y^{j,k}_{<t}$. Arithmetic intensity is very low as both the FLOPs and memory accesses scale linearly with the sequence length. With IPPD, the first decode step has $M_j$ queries. Writing $Z_{j,t}=\sum_{k'=1}^{M_j}(|x_{j,k'}|+|y^{j,k'}_{<t}|)$, FLOPs per token are increased by a factor of
\[
\frac{|p|+|c_j|+Z_{j,t}}
{|p|+|c_j|+|x_{j,k}|+|y^{j,k}_{<t}|}.
\]
By exposing reuse of $p$ and $c_j$ within a shared attention operation that generates $M_j$ tokens, IPPD reduces repeated memory traffic to the shared context. In subsequent decoding steps, some of the $M_j$ answers may complete generation, reducing the parallelism in future steps. We write $M_j^*$ in \Cref{tab:KV_size} to denote this effect. 

To conclude, the decode stage is characterized by a much lower arithmetic intensity than the prefill stage, resulting in a severe memory access bottleneck. Unlike prefill, the decode stage benefits from IPPD's memory access reduction even for extremely large $p$ and $c_j$. We see in \Cref{fig:results} that IPPD performs better than PC+PA for both datasets involving multi-token answers, which range from very short $p$ and $c_j$ in SQuAD 2.0 to very large $p$ and $c_j$ with 5-shot NarrativeQA.

\subsubsection{Comparison to Cascade Inference}
Cascade inference \citep{cascade-inference} fulfills a similar objective as IPPD: to reduce the number of memory accesses to the common contexts, thereby increasing arithmetic intensity. The way this is achieved is very different from IPPD: Cascade inference computes separate attention scores for keys that are part of the common prefix and those unique to a prompt. This allows queries from different prompts to attend to the common prefix keys in a shared operation, exposing reuse and reducing repeated memory traffic to those keys. Cascade inference also introduces more FLOPs, because a merge operator is required to combine the prefix attention scores with the unique suffixes for each prompt. For settings where the common prefix $p+c_j$ is much larger than the questions, IPPD's FLOPs increase is minimal, as shown in \Cref{app:prefill_efficiency}. In settings with long questions or generations relative to the context, the segmentation of attention by Cascade inference would be preferable. The real-world CCQA datasets used in this work more closely match the composition preferred by IPPD as evidenced in \cref{tab:dataset_info}, while general few-shot prompting is likely to contain longer suffixes and generations. Direct comparison with Cascade inference and other tree-based decoding methods mentioned in Section \ref{sec:related_work} is infeasible due to the lack of a general implementation compatible with the model families and inference platforms studied in this work, which are representative of real-world applications.

\section{Additional Experimental Details}
\label{app:exp_details}
\subsection{Evaluation Metric}

We evaluate candidate decoding strategies by benchmarking them against standard batched inference along two complementary dimensions: \textit{throughput} (efficiency) and \textit{answer quality} (fidelity to ground truth answers).

Throughput captures how many logical triplets can be processed per unit time per GPU, directly reflecting efficiency improvements. Concretely, let $|\mathcal{T}|$ denote the number of answered triplets, $T_{\mathrm{wall}}$ the measured wall-clock latency to produce all answers, and $G$ the number of GPUs used. We define
\begin{align}
    \mathrm{Throughput} \;=\; \frac{|\mathcal{T}|}{T_{\mathrm{wall}} \cdot G}.
\end{align}

Answer quality is measured at the task level using standard NLP evaluation metrics, including Accuracy, F1, and ROUGE-L. These metrics compare generated answers $\hat y_{j,k}$ to ground truth answers $y_{j,k}$ across all triplets. We use standard definitions for these metrics from the literature.

\subsection{Dataset Details} \label{app:datasets}

\paragraph{NarrativeQA \citep{kocisky2018_narrativeqa}} NarrativeQA is a reading comprehension benchmark containing books and movie scripts. We use the human-written summaries as document contexts. About 30 human-generated questions/answer pairs per document are provided. We report results on the test set using 5-shot prompting following the HELM benchmark \citep{liang2023holistic}.

\paragraph{SQuAD 2.0 \citep{rajpurkar2018_squadv2}} is an extractive reading comprehension task based on the original Stanford Question Answering Dataset. It contains human generated questions about Wikipedia articles, where the answer is a span of the input text. SQuAD 2.0 adds an additional set of unanswerable questions to increase the difficulty of the task. We report results on the publicly available development set, prompting the LLM to write "null" for unanswerable questions.

\paragraph{RACE \citep{lai2017_race}} RACE is an English reading comprehension dataset derived from  Chinese middle and high school exams. Multiple-choice questions are provided for each text passage by human English instructors covering a variety of topics. We report results on the test set, prompting the LLM to respond with the answer letter.

\paragraph{LongHealth \citep{adams2025_longhealth}} LongHealth is a collection of detailed patient cases each containing multiple discharge notes. Although realistic in form, structure and content, the documents are \textbf{entirely fictional}, so the dataset contains no real patient information. It was written by experienced physicians. We report results on the entire dataset, using the Task 1 setting from \citet{adams2025_longhealth}. We do not truncate documents, and modify their provided prompt to instruct the LLM to only respond with the answer letter.

\subsection{Inference Details} \label{app:inference_details}

\paragraph{General Details} We run all primary experiments on the Amazon EC2 g6e.8xlarge server using one Nvidia L40S 48GB GPU. The profiling and controlled ablations were run on a Nvidia A100 40GB GPU, which slightly varies the exact throughput metrics but retains the same overall results. We use greedy decoding for all generations, and non-thinking mode for all hybrid models. We also set a maximum output token length for each dataset: 40 for NarrativeQA, 30 for SQuAD 2.0 and 1 for RACE and LongHealth. Phi4-14B, Qwen3-32B and OLMo-2-32B are quantized to NF4 using BitsAndBytes. Tokenization time is excluded from time measurements. The time required to calculate the attention mask for IPPD is included.

\paragraph{Standard batched inference} We use the Hugging Face Transformers backend with the $generate()$ method and FlashAttention. Prefix caching is not enabled as it is not supported for batched inference. We increase the batch size for each model-dataset pair until throughput saturates or VRAM limits are exceeded.
\paragraph{Prefix caching + PagedAttention} We use the vLLM backend for optimized support of prefix caching and PagedAttention. Some vLLM features, such as the asynchronous V1 Engine, provide an inherent advantage over Transformers, irrespective of the inference acceleration methods used. To make a direct comparison between IPPD and prefix caching + PagedAttention, we adjust vLLM settings to exclude advantages that IPPD could also benefit from, were it also implemented on the same backend. We use the synchronous V0 Engine, enforcing eager mode as well as the use of the Transformers model implementation to match the one used by IPPD. We tune vLLM performance by providing access to all available VRAM, and setting the maximum model length to the longest prompt + output length for each dataset, so as to maximize the automatically managed batch size. FlashAttention is enabled.

\paragraph{IPPD} Our method is implemented with Hugging Face Transformers and Accelerate. The model code for each LLM is unchanged, as all IPPD functionality is achieved through a custom inference loop. We implement prefix caching only for $p$, the dataset-wide instruction. This affects NarrativeQA, where $p$ contains five few-shot examples. Prefix caching and IPPD are complementary in this setting: prefix caching avoids recomputing $p$, while IPPD reduces repeated processing of the contexts and questions within each stacked prompt. For each model-dataset pair, we select the optimal number of shared contexts per prompt and batch size. In practice, we find that these hyperparameters are largely independent of model size. The optimal number of contexts per prompt is largely determined by the context length, number of questions, and hardware architecture. \Cref{tab:hyperparameters} shows the hyperparameters used for each method. We largely keep the same IPPD hyperparameters across model size as far as VRAM limits allow.

\section{Additional Experimental Results} \label{app:add_results}
\subsection{Numerical Stability}
\label{app:numerical_stability}

IPPD follows the same idealized computational path as standard batched inference, so the remaining output variation is attributable to finite-precision arithmetic. Floating-point addition is not associative, and IPPD's stacked prompts differ in shape and size from a standard batch, which leads the GPU to select different tiling and reduction strategies when parallelizing the underlying matrix multiplications. Bitwise reproducibility is guaranteed only across runs with identically shaped inputs, a condition that batched inference and IPPD cannot satisfy simultaneously. These differences occasionally flip a greedy argmax and then propagate through the remainder of an answer, so longer generations accumulate more divergence. They introduce no systematic change in task-level quality.

\subsection{Throughput Measurements}
Table \ref{tab:add_results} presents absolute throughput scores, i.e., questions answered per second (QPS), for all benchmarks.

\begin{table*}[t!]
\centering\small
\aboverulesep=0ex
\belowrulesep=0.25ex

\begin{tabular}{@{}llcccccccc@{}}
\toprule
\multirow{3}{*}{Model}      &                          & \multicolumn{2}{c}{\multirow{2}{*}{NarrativeQA (5s)}} & \multicolumn{2}{c}{\multirow{2}{*}{SQuAD 2.0}} & \multicolumn{2}{c}{\multirow{2}{*}{RACE}} & \multicolumn{2}{c}{\multirow{2}{*}{LongHealth}}                                       \\
   &  & \multicolumn{2}{c}{}                         & \multicolumn{2}{c}{}                         & \multicolumn{2}{c}{}                         & \multicolumn{2}{c}{}                         \\
 & & B    & C/P                       & B    & C/P                         & B    & C/P                       & B    & C/P        \\
                            \midrule                                   \\
\multirow{6}{*}{Batched Inf.} & \multicolumn{1}{l|}{Qwen3-32B}  & 8      & \multicolumn{1}{c|}{1}      &   30    & \multicolumn{1}{c|}{1}      &  5    & \multicolumn{1}{c|}{1}      &   1    &   1         \\
                    & \multicolumn{1}{l|}{OLMo-2-32B} & 8      & \multicolumn{1}{c|}{1}      &   30    & \multicolumn{1}{c|}{1}      &   5    & \multicolumn{1}{c|}{1}      &   1    &     1      \\ 
                    & \multicolumn{1}{l|}{Phi-4-14B}  & 8      & \multicolumn{1}{c|}{1}      &   30    & \multicolumn{1}{c|}{1}      &  5    & \multicolumn{1}{c|}{1}      &   1    &   1         \\
                    & \multicolumn{1}{l|}{Qwen3-8B} & 20      & \multicolumn{1}{c|}{1}      &   30    & \multicolumn{1}{c|}{1}      &   5    & \multicolumn{1}{c|}{1}      &   1    &     1      \\      
                    & \multicolumn{1}{l|}{Qwen3-4B-Instr.}  & 20      & \multicolumn{1}{c|}{1}      &   30    & \multicolumn{1}{c|}{1}      &  5    & \multicolumn{1}{c|}{1}      &  1    &   1         \\
                    & \multicolumn{1}{l|}{Qwen3-1.7B} & 20      & \multicolumn{1}{c|}{1}      &   30    & \multicolumn{1}{c|}{1}      &   5    & \multicolumn{1}{c|}{1}      &   1    &     1      \\ \midrule   
\multirow{6}{*}{IPPD}& \multicolumn{1}{l|}{Qwen3-32B}  & 2      & \multicolumn{1}{c|}{3}      &   5    & \multicolumn{1}{c|}{6}      &  2    & \multicolumn{1}{c|}{2}      &   1    &   1         \\
                    & \multicolumn{1}{l|}{OLMo-2-32B} & 2      & \multicolumn{1}{c|}{3}      &   5    & \multicolumn{1}{c|}{6}      &   2    & \multicolumn{1}{c|}{2}      &   1    &     1      \\ 
                    & \multicolumn{1}{l|}{Phi-4-14B}  & 3      & \multicolumn{1}{c|}{4}      &   6    & \multicolumn{1}{c|}{6}      &  2    & \multicolumn{1}{c|}{2}      &   1    &   1         \\
                    & \multicolumn{1}{l|}{Qwen3-8B} & 3      & \multicolumn{1}{c|}{4}      &   6    & \multicolumn{1}{c|}{6}      &   2    & \multicolumn{1}{c|}{2}      &   1    &     1      \\     
                    & \multicolumn{1}{l|}{Qwen3-4B-Instr.}  & 3      & \multicolumn{1}{c|}{4} &   6    & \multicolumn{1}{c|}{6}      &  2    & \multicolumn{1}{c|}{2}      &   1    &   1         \\
                    & \multicolumn{1}{l|}{Qwen3-1.7B} & 3      & \multicolumn{1}{c|}{4}      &   6    & \multicolumn{1}{c|}{6}      &   2    & \multicolumn{1}{c|}{2}      &   1    &     1      \\   
\bottomrule
                           
\end{tabular}
\caption{Hyperparameter selection for batch size B and contexts stacked per prompt C/P for IPPD and standard batched inference. Hyperparameters are mostly constant across model sizes, except when VRAM limits require a lower batch size. }
\label{tab:hyperparameters}
\end{table*}

\begin{table*}[ht!]
\centering\small
\aboverulesep=0ex
\belowrulesep=0.25ex

\begin{tabular}{@{}llcccc@{}}
\toprule
\multirow{2}{*}{Model}      &                          & NarrativeQA (5s) & LongHealth & SQuAD 2.0 & RACE \\
                            &                          & QPS              & QPS        & QPS       & QPS  \\
                            \midrule

\multirow{3}{*}{Qwen3-32B}  & \multicolumn{1}{l|}{Batched Inference}  & 0.48 & 0.18 & 2.86  & 4.52 \\
                            & \multicolumn{1}{l|}{PC + PA} & 7.06   & \textbf{2.82} & 11.90  & 5.84 \\
                            & \multicolumn{1}{l|}{IPPD} & \textbf{10.74}           & 1.93 & \textbf{15.74} & \textbf{11.78} \\ \midrule

\multirow{3}{*}{OLMo-2-32B} & \multicolumn{1}{l|}{Batched Inference}  & 0.54 & - & 3.47  & 4.75 \\
                            & \multicolumn{1}{l|}{PC + PA} & 7.15   & - & 12.54 & 6.15 \\
                            & \multicolumn{1}{l|}{IPPD} & \textbf{14.53}           & - & \textbf{19.43} & \textbf{12.91} \\ \midrule

\multirow{3}{*}{Phi4-14B}   & \multicolumn{1}{l|}{Batched Inference}  & 0.91 & 0.44 & 5.36  & 10.84 \\
                            & \multicolumn{1}{l|}{PC + PA} & 14.42  & \textbf{7.02} & 27.11 & 13.90 \\
                            & \multicolumn{1}{l|}{IPPD} & \textbf{26.13}           & 5.05 & \textbf{38.55} & \textbf{28.55} \\ \midrule

\multirow{3}{*}{Qwen3-8B}   & \multicolumn{1}{l|}{Batched Inference}  & 1.75 & 0.71  & 14.25 & 21.18 \\
                            & \multicolumn{1}{l|}{PC + PA} & 26.39  & \textbf{11.43} & 50.10  & 25.17 \\
                            & \multicolumn{1}{l|}{IPPD} & \textbf{57.77}           & 6.95  & \textbf{62.04} & \textbf{52.12} \\ \midrule

\multirow{3}{*}{Qwen3-4B-Instr.} & \multicolumn{1}{l|}{Batched Inference}  & 1.90  & 1.02  & 18.01 & 34.72 \\
                            & \multicolumn{1}{l|}{PC + PA} & 37.84 & \textbf{16.00}    & 79.68 & 40.11 \\
                            & \multicolumn{1}{l|}{IPPD} & \textbf{54.48}           & 8.92  & \textbf{82.88} & \textbf{82.98} \\ \midrule

\multirow{3}{*}{Qwen3-1.7B} & \multicolumn{1}{l|}{Batched Inference}  & 4.54 & 2.40   & 35.36  & 77.74 \\
                            & \multicolumn{1}{l|}{PC + PA} & 78.20   & \textbf{33.33} & \textbf{172.07} & 94.88 \\
                            & \multicolumn{1}{l|}{IPPD} & \textbf{105.91}          & 20.34 & 145.02 & \textbf{169.79} \\ \bottomrule
\end{tabular}
\caption{Throughput comparison of all models on four selected CCQA datasets. We compare batched inference, prefix caching + PagedAttention (PC + PA), and our proposed IPPD. We use queries per second (QPS) as the measurement. The highest score for each setting is in \textbf{bold}.
}
\label{tab:add_results}
\end{table*}

\section{Profiling and Controlled Ablations}
\label{app:camera_ready_experiments}

\newcommand{\cameraPrefillDecodeTable}{%
\begin{table*}[t!]
\centering\small
\aboverulesep=0ex
\belowrulesep=0.25ex
\begin{tabular}{@{}llcccc@{}}
\toprule
Model & Dataset & Prefill steps & Prefill time & Decode steps & Decode time \\
\midrule
\multirow{4}{*}{Qwen3-32B} & NarrativeQA (5s) & 20.95$\times$ & 37.59$\times$ & 10.25$\times$ & 8.66$\times$ \\
 & SQuAD 2.0 & 7.07$\times$ & 5.00$\times$ & 5.79$\times$ & 5.08$\times$ \\
 & RACE & 2.80$\times$ & 2.58$\times$ & -- & -- \\
 & LongHealth & 20.00$\times$ & 9.33$\times$ & -- & -- \\
\midrule
\multirow{4}{*}{OLMo-2-32B} & NarrativeQA (5s) & 20.95$\times$ & 43.68$\times$ & 10.45$\times$ & 11.81$\times$ \\
 & SQuAD 2.0 & 6.19$\times$ & 5.86$\times$ & 4.02$\times$ & 4.18$\times$ \\
 & RACE & 2.80$\times$ & 2.67$\times$ & -- & -- \\
 & LongHealth & -- & -- & -- & -- \\
\midrule
\multirow{4}{*}{Phi-4-14B} & NarrativeQA (5s) & 41.25$\times$ & 37.92$\times$ & 23.16$\times$ & 17.86$\times$ \\
 & SQuAD 2.0 & 9.90$\times$ & 5.42$\times$ & 9.64$\times$ & 8.49$\times$ \\
 & RACE & 2.80$\times$ & 2.71$\times$ & -- & -- \\
 & LongHealth & 20.00$\times$ & 9.99$\times$ & -- & -- \\
\midrule
\multirow{4}{*}{Qwen3-8B} & NarrativeQA (5s) & 22.00$\times$ & 29.60$\times$ & 11.77$\times$ & 10.72$\times$ \\
 & SQuAD 2.0 & 9.90$\times$ & 4.40$\times$ & 7.73$\times$ & 6.35$\times$ \\
 & RACE & 2.80$\times$ & 2.48$\times$ & -- & -- \\
 & LongHealth & 20.00$\times$ & 8.26$\times$ & -- & -- \\
\midrule
\multirow{4}{*}{Qwen3-4B-Instr.} & NarrativeQA (5s) & 16.50$\times$ & 26.72$\times$ & 9.48$\times$ & 10.88$\times$ \\
 & SQuAD 2.0 & 9.90$\times$ & 4.35$\times$ & 8.26$\times$ & 7.61$\times$ \\
 & RACE & 2.80$\times$ & 2.44$\times$ & -- & -- \\
 & LongHealth & 20.00$\times$ & 6.97$\times$ & -- & -- \\
\midrule
\multirow{4}{*}{Qwen3-1.7B} & NarrativeQA (5s) & 16.50$\times$ & 23.72$\times$ & 8.76$\times$ & 12.15$\times$ \\
 & SQuAD 2.0 & 9.90$\times$ & 3.88$\times$ & 6.65$\times$ & 7.52$\times$ \\
 & RACE & 2.80$\times$ & 2.45$\times$ & -- & -- \\
 & LongHealth & 20.00$\times$ & 6.67$\times$ & -- & -- \\
\bottomrule
\end{tabular}
\caption{Prefill and decode step and time reductions from IPPD relative to batched autoregressive inference. Values are multiplicative speedups (higher is better). Dashes denote datasets with single-token outputs or configurations that exceed the model's maximum sequence length.}
\label{tab:camera-prefill-decode}
\end{table*}
}

\newcommand{\cameraQuestionsTable}{%
\begin{table*}[t!]
\centering\small
\aboverulesep=0ex
\belowrulesep=0.25ex
\resizebox{\textwidth}{!}{%
\begin{tabular}{@{}l@{\hspace{8pt}}c@{\hspace{2pt}}c@{\hspace{10pt}}c@{\hspace{2pt}}c@{\hspace{10pt}}c@{\hspace{2pt}}c@{\hspace{10pt}}c@{\hspace{2pt}}c@{\hspace{10pt}}c@{\hspace{2pt}}c@{\hspace{10pt}}c@{\hspace{2pt}}c@{}}
\toprule
\multirow{3}{*}{Model} & \multicolumn{12}{c}{Questions / context} \\
 & \multicolumn{2}{c}{1} & \multicolumn{2}{c}{2} & \multicolumn{2}{c}{4} & \multicolumn{2}{c}{8} & \multicolumn{2}{c}{16} & \multicolumn{2}{c}{32} \\
 & PC+PA & IPPD & PC+PA & IPPD & PC+PA & IPPD & PC+PA & IPPD & PC+PA & IPPD & PC+PA & IPPD \\
\midrule
\multicolumn{13}{l}{\textit{NarrativeQA (5s)}} \\
Qwen3-32B & \textbf{5.09$\times$} & 1.52$\times$ & \textbf{5.72$\times$} & 2.63$\times$ & \textbf{6.73$\times$} & 4.34$\times$ & \textbf{7.93$\times$} & 7.20$\times$ & 10.81$\times$ & \textbf{12.28$\times$} & 15.76$\times$ & \textbf{18.91$\times$} \\
OLMo-2-32B & -- & -- & -- & -- & -- & -- & -- & -- & -- & -- & -- & -- \\
Phi-4-14B & \textbf{6.50$\times$} & 1.57$\times$ & \textbf{7.07$\times$} & 2.74$\times$ & \textbf{8.38$\times$} & 5.02$\times$ & \textbf{9.77$\times$} & 8.84$\times$ & 12.33$\times$ & \textbf{15.60$\times$} & 16.60$\times$ & \textbf{23.98$\times$} \\
Qwen3-8B & \textbf{5.08$\times$} & 1.55$\times$ & \textbf{5.17$\times$} & 2.77$\times$ & \textbf{5.67$\times$} & 5.11$\times$ & 6.71$\times$ & \textbf{8.83$\times$} & 7.95$\times$ & \textbf{14.01$\times$} & 11.51$\times$ & \textbf{20.59$\times$} \\
Qwen3-4B-Instr. & \textbf{5.70$\times$} & 1.26$\times$ & \textbf{6.01$\times$} & 2.21$\times$ & \textbf{6.43$\times$} & 3.90$\times$ & \textbf{7.57$\times$} & 6.51$\times$ & 9.19$\times$ & \textbf{10.64$\times$} & 12.53$\times$ & \textbf{15.86$\times$} \\
Qwen3-1.7B & \textbf{5.90$\times$} & 1.19$\times$ & \textbf{5.99$\times$} & 2.12$\times$ & \textbf{6.43$\times$} & 3.56$\times$ & \textbf{7.46$\times$} & 6.10$\times$ & 8.56$\times$ & \textbf{9.44$\times$} & 11.51$\times$ & \textbf{13.97$\times$} \\
\midrule
\multicolumn{13}{l}{\textit{SQuAD 2.0}} \\
Qwen3-32B & \textbf{4.11$\times$} & 1.10$\times$ & \textbf{4.27$\times$} & 1.83$\times$ & \textbf{4.41$\times$} & 3.15$\times$ & 4.56$\times$ & \textbf{4.94$\times$} & 4.62$\times$ & \textbf{5.61$\times$} & 4.77$\times$ & \textbf{6.04$\times$} \\
OLMo-2-32B & \textbf{4.09$\times$} & 1.23$\times$ & \textbf{4.08$\times$} & 1.86$\times$ & \textbf{4.00$\times$} & 3.16$\times$ & 3.96$\times$ & \textbf{4.03$\times$} & 3.83$\times$ & \textbf{5.84$\times$} & 3.77$\times$ & \textbf{6.17$\times$} \\
Phi-4-14B & \textbf{5.24$\times$} & 1.18$\times$ & \textbf{5.32$\times$} & 2.10$\times$ & \textbf{5.31$\times$} & 3.60$\times$ & 5.48$\times$ & \textbf{5.95$\times$} & 5.69$\times$ & \textbf{8.49$\times$} & 6.20$\times$ & \textbf{9.70$\times$} \\
Qwen3-8B & \textbf{4.18$\times$} & 1.17$\times$ & \textbf{4.22$\times$} & 1.85$\times$ & \textbf{4.18$\times$} & 2.97$\times$ & 4.06$\times$ & \textbf{4.29$\times$} & 4.16$\times$ & \textbf{5.46$\times$} & 4.28$\times$ & \textbf{5.50$\times$} \\
Qwen3-4B-Instr. & \textbf{5.80$\times$} & 1.11$\times$ & \textbf{6.11$\times$} & 1.96$\times$ & \textbf{5.76$\times$} & 3.13$\times$ & \textbf{5.99$\times$} & 4.88$\times$ & 5.84$\times$ & \textbf{6.21$\times$} & 6.10$\times$ & \textbf{6.59$\times$} \\
Qwen3-1.7B & \textbf{6.91$\times$} & 1.07$\times$ & \textbf{6.67$\times$} & 1.73$\times$ & \textbf{6.61$\times$} & 2.73$\times$ & \textbf{6.69$\times$} & 4.30$\times$ & \textbf{6.37$\times$} & 5.54$\times$ & \textbf{6.57$\times$} & 6.07$\times$ \\
\midrule
\multicolumn{13}{l}{\textit{RACE}} \\
Qwen3-32B & \textbf{1.56$\times$} & 1.05$\times$ & 1.50$\times$ & \textbf{1.80$\times$} & 1.49$\times$ & \textbf{2.77$\times$} & 1.52$\times$ & \textbf{3.90$\times$} & 1.79$\times$ & \textbf{4.63$\times$} & 1.97$\times$ & \textbf{4.84$\times$} \\
OLMo-2-32B & \textbf{1.55$\times$} & 1.08$\times$ & 1.50$\times$ & \textbf{1.83$\times$} & 1.51$\times$ & \textbf{2.94$\times$} & 1.55$\times$ & \textbf{4.16$\times$} & 1.78$\times$ & \textbf{5.19$\times$} & 1.93$\times$ & \textbf{5.28$\times$} \\
Phi-4-14B & \textbf{1.58$\times$} & 1.05$\times$ & 1.53$\times$ & \textbf{1.76$\times$} & 1.51$\times$ & \textbf{2.82$\times$} & 1.53$\times$ & \textbf{3.97$\times$} & 1.83$\times$ & \textbf{5.07$\times$} & 1.96$\times$ & \textbf{5.35$\times$} \\
Qwen3-8B & \textbf{1.59$\times$} & 1.03$\times$ & 1.51$\times$ & \textbf{1.69$\times$} & 1.49$\times$ & \textbf{2.58$\times$} & 1.48$\times$ & \textbf{3.35$\times$} & 1.75$\times$ & \textbf{3.97$\times$} & 1.93$\times$ & \textbf{4.04$\times$} \\
Qwen3-4B-Instr. & \textbf{1.71$\times$} & 1.00$\times$ & 1.61$\times$ & \textbf{1.65$\times$} & 1.53$\times$ & \textbf{2.42$\times$} & 1.55$\times$ & \textbf{3.25$\times$} & 1.84$\times$ & \textbf{3.81$\times$} & 1.93$\times$ & \textbf{3.70$\times$} \\
Qwen3-1.7B & \textbf{1.95$\times$} & 0.94$\times$ & \textbf{1.79$\times$} & 1.54$\times$ & 1.72$\times$ & \textbf{2.29$\times$} & 1.72$\times$ & \textbf{3.08$\times$} & 2.03$\times$ & \textbf{3.71$\times$} & 2.19$\times$ & \textbf{3.70$\times$} \\
\midrule
\multicolumn{13}{l}{\textit{LongHealth}} \\
Qwen3-32B & \textbf{0.97$\times$} & 0.58$\times$ & \textbf{1.92$\times$} & 1.16$\times$ & \textbf{3.71$\times$} & 2.23$\times$ & \textbf{7.16$\times$} & 4.30$\times$ & \textbf{12.80$\times$} & 7.81$\times$ & \textbf{21.81$\times$} & 13.13$\times$ \\
OLMo-2-32B & -- & -- & -- & -- & -- & -- & -- & -- & -- & -- & -- & -- \\
Phi-4-14B & \textbf{1.06$\times$} & 0.65$\times$ & \textbf{1.95$\times$} & 1.20$\times$ & \textbf{3.74$\times$} & 2.34$\times$ & \textbf{7.15$\times$} & 4.41$\times$ & \textbf{12.91$\times$} & 7.98$\times$ & \textbf{21.53$\times$} & 13.68$\times$ \\
Qwen3-8B & \textbf{1.13$\times$} & 0.58$\times$ & \textbf{2.03$\times$} & 1.06$\times$ & \textbf{3.92$\times$} & 2.00$\times$ & \textbf{6.92$\times$} & 3.75$\times$ & \textbf{12.47$\times$} & 6.83$\times$ & \textbf{20.02$\times$} & 11.44$\times$ \\
Qwen3-4B-Instr. & \textbf{1.21$\times$} & 0.51$\times$ & \textbf{2.06$\times$} & 0.89$\times$ & \textbf{3.72$\times$} & 1.68$\times$ & \textbf{6.83$\times$} & 3.16$\times$ & \textbf{11.86$\times$} & 5.64$\times$ & \textbf{18.48$\times$} & 9.31$\times$ \\
Qwen3-1.7B & \textbf{1.48$\times$} & 0.58$\times$ & \textbf{2.33$\times$} & 0.90$\times$ & \textbf{4.05$\times$} & 1.65$\times$ & \textbf{6.76$\times$} & 2.95$\times$ & \textbf{10.56$\times$} & 5.12$\times$ & \textbf{15.80$\times$} & 8.63$\times$ \\
\midrule
\textbf{Total wins} & \textbf{22} & 0 & \textbf{17} & 5 & \textbf{16} & 6 & \textbf{11} & \textbf{11} & 6 & \textbf{16} & 6 & \textbf{16} \\
\bottomrule
\end{tabular}%
}
\caption{Questions-per-context ablation across datasets and models. Values are throughput speedups relative to batched autoregressive inference (higher is better). The better method for each model and column is in \textbf{bold}, and dashes denote configurations that exceed the model's maximum sequence length.}
\label{tab:camera-questions}
\end{table*}
}

\newcommand{\cameraContextTable}{%
\begin{table*}[t!]
\centering\small
\aboverulesep=0ex
\belowrulesep=0.25ex
\resizebox{0.73\textwidth}{!}{%
\begin{tabular}{@{}l@{\hspace{8pt}}c@{\hspace{2pt}}c@{\hspace{10pt}}c@{\hspace{2pt}}c@{\hspace{10pt}}c@{\hspace{2pt}}c@{\hspace{10pt}}c@{\hspace{2pt}}c@{}}
\toprule
\multirow{3}{*}{Model} & \multicolumn{8}{c}{Context length} \\
 & \multicolumn{2}{c}{1$\times$} & \multicolumn{2}{c}{2$\times$} & \multicolumn{2}{c}{4$\times$} & \multicolumn{2}{c}{8$\times$} \\
 & PC+PA & IPPD & PC+PA & IPPD & PC+PA & IPPD & PC+PA & IPPD \\
\midrule
\multicolumn{9}{l}{\textit{NarrativeQA (5s)}} \\
Qwen3-32B & 15.97$\times$ & \textbf{18.25$\times$} & \textbf{20.50$\times$} & 17.73$\times$ & \textbf{20.74$\times$} & 13.44$\times$ & \textbf{18.74$\times$} & 12.91$\times$ \\
OLMo-2-32B & 14.39$\times$ & \textbf{22.07$\times$} & -- & -- & -- & -- & -- & -- \\
Phi-4-14B & 16.05$\times$ & \textbf{23.16$\times$} & \textbf{21.46$\times$} & 18.85$\times$ & -- & -- & -- & -- \\
Qwen3-8B & 10.91$\times$ & \textbf{19.72$\times$} & 13.54$\times$ & \textbf{16.66$\times$} & 12.02$\times$ & \textbf{14.39$\times$} & 12.10$\times$ & \textbf{13.55$\times$} \\
Qwen3-4B-Instr. & 12.13$\times$ & \textbf{15.02$\times$} & \textbf{15.07$\times$} & 13.08$\times$ & \textbf{13.83$\times$} & 10.76$\times$ & \textbf{12.73$\times$} & 9.14$\times$ \\
Qwen3-1.7B & 11.02$\times$ & \textbf{13.35$\times$} & \textbf{12.56$\times$} & 11.78$\times$ & \textbf{11.58$\times$} & 10.21$\times$ & \textbf{10.43$\times$} & 8.34$\times$ \\
\midrule
\multicolumn{9}{l}{\textit{SQuAD 2.0}} \\
Qwen3-32B & 4.64$\times$ & \textbf{4.86$\times$} & 3.52$\times$ & \textbf{4.77$\times$} & 3.56$\times$ & \textbf{4.76$\times$} & \textbf{3.99$\times$} & 3.54$\times$ \\
OLMo-2-32B & 3.87$\times$ & \textbf{4.51$\times$} & 3.02$\times$ & \textbf{5.07$\times$} & 2.81$\times$ & \textbf{5.17$\times$} & 3.11$\times$ & \textbf{4.13$\times$} \\
Phi-4-14B & 5.52$\times$ & \textbf{6.69$\times$} & 4.03$\times$ & \textbf{6.55$\times$} & 4.03$\times$ & \textbf{7.33$\times$} & 5.05$\times$ & \textbf{8.06$\times$} \\
Qwen3-8B & 4.12$\times$ & \textbf{4.66$\times$} & 3.14$\times$ & \textbf{4.55$\times$} & 2.86$\times$ & \textbf{4.38$\times$} & 2.98$\times$ & \textbf{4.31$\times$} \\
Qwen3-4B-Instr. & \textbf{5.83$\times$} & 5.18$\times$ & 4.34$\times$ & \textbf{5.17$\times$} & 3.57$\times$ & \textbf{4.57$\times$} & 3.44$\times$ & \textbf{4.02$\times$} \\
Qwen3-1.7B & \textbf{6.64$\times$} & 4.61$\times$ & \textbf{4.70$\times$} & 4.48$\times$ & 3.77$\times$ & \textbf{4.25$\times$} & 3.24$\times$ & \textbf{3.78$\times$} \\
\midrule
\multicolumn{9}{l}{\textit{RACE}} \\
Qwen3-32B & 1.50$\times$ & \textbf{2.54$\times$} & 1.37$\times$ & \textbf{2.68$\times$} & 1.32$\times$ & \textbf{2.55$\times$} & 1.36$\times$ & \textbf{2.28$\times$} \\
OLMo-2-32B & 1.50$\times$ & \textbf{2.63$\times$} & 1.34$\times$ & \textbf{2.80$\times$} & 1.29$\times$ & \textbf{2.71$\times$} & 1.35$\times$ & \textbf{2.78$\times$} \\
Phi-4-14B & 1.54$\times$ & \textbf{2.62$\times$} & 1.36$\times$ & \textbf{2.70$\times$} & 1.33$\times$ & \textbf{2.64$\times$} & 1.36$\times$ & \textbf{2.40$\times$} \\
Qwen3-8B & 1.49$\times$ & \textbf{2.34$\times$} & 1.40$\times$ & \textbf{2.46$\times$} & 1.38$\times$ & \textbf{2.41$\times$} & 1.42$\times$ & \textbf{2.12$\times$} \\
Qwen3-4B-Instr. & 1.57$\times$ & \textbf{2.25$\times$} & 1.44$\times$ & \textbf{2.30$\times$} & 1.41$\times$ & \textbf{2.17$\times$} & 1.40$\times$ & \textbf{1.81$\times$} \\
Qwen3-1.7B & 1.72$\times$ & \textbf{2.10$\times$} & 1.51$\times$ & \textbf{2.18$\times$} & 1.41$\times$ & \textbf{2.10$\times$} & 1.36$\times$ & \textbf{1.78$\times$} \\
\midrule
\textbf{Total wins} & 2 & \textbf{16} & 5 & \textbf{12} & 3 & \textbf{13} & 4 & \textbf{12} \\
\bottomrule
\end{tabular}%
}
\caption{Context-length ablation across datasets and models. Values are throughput speedups relative to batched autoregressive inference (higher is better). The better method for each model and column is in \textbf{bold}, and dashes denote configurations that exceed the model's maximum sequence length.}
\label{tab:camera-context}
\end{table*}
}

\subsection{Prefill and Decode Speedup}
\label{app:prefill_decode}

\cameraPrefillDecodeTable

\paragraph{Purpose and methodology.}
We separately profile the prefill and decode phases to determine whether IPPD's end-to-end throughput gains arise in both phases of inference. For standard batched autoregressive inference and IPPD, we record the total number of prefill and decode steps and their cumulative wall-clock time. We exclude PC+PA from this phase-level comparison because its vLLM scheduler uses a different execution pattern, scheduling consecutive prefills and large decode batches backed by paged KV-cache entries. This makes their separate timings difficult to compare directly with the Transformers backend. \Cref{tab:camera-prefill-decode} reports the ratio between the autoregressive and IPPD measurements, such that values above $1\times$ indicate a reduction from IPPD. We use the original, unmodified datasets and otherwise follow the general setup in \Cref{app:inference_details}, maximizing the batch size for each setting subject to the available VRAM.

\paragraph{Results.}
IPPD accelerates both prefill and decode relative to standard batched autoregressive inference. During prefill, IPPD reduces the number of forward passes by jointly processing questions that share a context and by stacking multiple contexts in a prompt. Although each IPPD prefill step processes a larger sequence and therefore requires more FLOPs than a corresponding batched-inference step, the reduction in step count yields a substantial reduction in total prefill time. In most settings, the time reduction is smaller than the step reduction because prefill already has relatively high arithmetic intensity, making its additional FLOPs more consequential. The decode-step reduction is smaller because the number of steps is determined by the longest active answer in each batch. Nevertheless, decode-time reductions mostly track the corresponding step reductions. Thus, the additional FLOPs per IPPD decode step do not substantially increase its latency, consistent with decode being primarily memory-bandwidth bound as analyzed in \Cref{app:comp_efficiency_details}.

\subsection{Controlled Ablations}
\label{app:controlled_ablations}

\cameraQuestionsTable

\paragraph{Controlled-ablation design.}
For the following two ablations, we perturb real benchmark examples rather than construct fully synthetic prompts. In practice, throughput depends on the joint distribution of instruction, context, question, and answer lengths, which varies across examples and datasets. A synthetic sweep that varies one length while holding the others fixed would not capture these interactions and may not represent realistic workloads. Our controlled variants therefore modify one axis of the original datasets while retaining their remaining content and natural length variation.

\subsubsection{Question Count Ablation}
\label{app:question_ablation}

\paragraph{Purpose and methodology.}
This ablation measures how IPPD scales with the amount of parallelism available among questions that share a context. For each dataset, we construct variants containing exactly 1, 2, 4, 8, 16, or 32 questions per context. For a target count $M$, we retain the first $M$ question--answer pairs when a context contains at least $M$ pairs. When it contains fewer than $M$, we cycle through its existing pairs until the target count is reached, duplicating the corresponding answers and options and numbering repeated query strings so that they remain distinct inputs. The context and other prompt components remain fixed. We evaluate IPPD and PC+PA on each variant and report throughput relative to standard batched autoregressive inference in \Cref{tab:camera-questions}. All other settings follow \Cref{app:inference_details}.

\paragraph{Results.}
The speedup from IPPD depends strongly on the number of questions sharing a context. With one question per context, shared-context parallelism is absent, and IPPD provides little or no benefit beyond that obtained by stacking distinct contexts in one prompt. Its speedup then grows rapidly with the question count and is generally largest at 32 questions per context. PC+PA provides a substantial initial gain at one question per context, but scales more slowly as additional questions are added. Consequently, IPPD overtakes PC+PA in an increasing number of model--dataset settings. LongHealth is the main exception. Its contexts are extremely long and its answers contain only one token, leaving no iterative decode phase. PC+PA is therefore consistently faster than IPPD by a broadly similar factor as the question count increases. We attribute much of this difference to the efficient FlashAttention kernels used by the PC+PA backend, whereas IPPD uses standard scaled dot-product attention (SDPA) to support its custom attention mask.

\cameraContextTable

\subsubsection{Context Length Ablation}
\label{app:context_ablation}

\paragraph{Purpose and methodology.}
This ablation evaluates how increasing the shared-context length affects the relative throughput of IPPD and PC+PA. We exclude LongHealth due to its existing long context. For the remaining three datasets, we retain each example's questions and construct nested context variants by concatenating contexts sampled from other randomly drawn examples, producing context-length factors of $1\times$, $2\times$, $4\times$, and $8\times$ on average. We rerun all three inference methods for each variant and report speedup relative to standard batched autoregressive inference in \Cref{tab:camera-context}. The remaining setup follows \Cref{app:inference_details}.

\paragraph{Results.}
Increasing context length generally reduces IPPD's speedup relative to autoregressive inference, although the trend is not monotonic for every model. As the context grows, a larger fraction of inference time is spent in the FLOP-intensive prefill phase, reducing the relative benefit of IPPD's memory-access savings. Even at $8\times$ context length, however, IPPD retains substantial speedups on NarrativeQA, SQuAD 2.0, and RACE. It outperforms PC+PA in 12 of the 16 supported model--dataset comparisons at this context length, demonstrating that its benefit persists well beyond the original context-length distribution.

\end{document}